\documentclass[conference]{IEEEtran}
\usepackage{cite}
\usepackage{amsmath,amssymb,amsfonts}
\usepackage{algorithmic}
\usepackage{graphicx}
\usepackage{textcomp}
\usepackage{xcolor}
\usepackage{url}
\usepackage{hyperref}

\usepackage{amsthm}
\usepackage[shortlabels]{enumitem}
\usepackage{caption}
\usepackage{array,multirow}

\theoremstyle{plain}
\newtheorem{theorem}{Theorem}
\newtheorem{corollary}[theorem]{Corollary}
\newtheorem{lemma}[theorem]{Lemma}

\theoremstyle{definition}

\newtheorem{definition}[theorem]{Definition}

\def\BibTeX{{\rm B\kern-.05em{\sc i\kern-.025em b}\kern-.08em
    T\kern-.1667em\lower.7ex\hbox{E}\kern-.125emX}}
\begin{document}

\title{Ensemble Learning using Error Correcting Output Codes: New Classification Error Bounds}

\author{\IEEEauthorblockN{Hieu D. Nguyen\IEEEauthorrefmark{1}}
\IEEEauthorblockA{nguyen@rowan.edu}
\and
\IEEEauthorblockN{Mohammed Sarosh Khan\IEEEauthorrefmark{2}}
\IEEEauthorblockA{khanmo67@students.rowan.edu}
\and
\IEEEauthorblockN{Nicholas Kaegi\IEEEauthorrefmark{2}}
\IEEEauthorblockA{kaegin63@students.rowan.edu}
\and
\IEEEauthorblockN{Shen-Shyang Ho\IEEEauthorrefmark{2}}
\IEEEauthorblockA{hos@rowan.edu}
\and
\IEEEauthorblockN{Jonathan Moore\IEEEauthorrefmark{1}}
\IEEEauthorblockA{moorej9@students.rowan.edu}
\and
\IEEEauthorblockN{Logan Borys\IEEEauthorrefmark{1}}
\IEEEauthorblockA{borysl3@students.rowan.edu}
\and
\IEEEauthorblockN{Lucas Lavalva\IEEEauthorrefmark{2}}
\IEEEauthorblockA{lavalv46@students.rowan.edu}
\\
\and
\IEEEauthorblockA{\IEEEauthorrefmark{1}\textit{Department of Mathematics},
\textit{Rowan University},
Glassboro, NJ USA.}
\IEEEauthorblockA{\IEEEauthorrefmark{2}\textit{Department of Computer Science},
\textit{Rowan University},
Glassboro, NJ USA.}
}

\maketitle

\begin{abstract}
New bounds on classification error rates for the error-correcting output code (ECOC) approach in machine learning are presented.  These bounds have exponential decay complexity with respect to codeword length and theoretically validate the effectiveness of the ECOC approach.  Bounds are derived for two different models: the first under the assumption that all base classifiers are independent and the second under the assumption that all base classifiers are mutually correlated up to first-order.  Moreover, we perform ECOC classification on six datasets and compare their error rates with our bounds to experimentally validate our work and show the effect of correlation on classification accuracy.
\end{abstract}


\section{Introduction}
Error correcting output codes (ECOC) is an ensemble classification technique in machine learning that is motivated by coding theory where transmitted or stored information is encoded by binary strings (codewords) with high Hamming distance which allows for unique decoding of bit errors~\cite{db}. There are many variants and extension of the ECOC techniques such as 
the use of ternary~\cite{escalera2008decoding} and $N$-ary codes~\cite{zhou2019}, optimizing individual classifier performance concurrently by exploiting their relationships~\cite{liu2015joint}, and optimizing the learning of the base classifiers together as a   multi-task learning problem~\cite{zhong2013adaptive}. 
Some theoretical error bounds for ECOC can be found in ~\cite{db} and~\cite{zhou2019}. Moreover, Passerini et al.~\cite{passerini2004new} provided leave-one-out error bound for using kernel machines as base classifiers for ECOC classifier. More recently, the ECOC technique has been extended to handle the zero-shot learning problem~\cite{qin2017zero}, the life-long learning problem~\cite{ho2020error}, and handling adversarial examples in neural network by integrating ECOC with increasing ensemble diversity~\cite{song2021error}.

In a conventional ECOC classifier, each class of a given dataset is assigned a codeword and a learned model $L$ is trained through an ensemble of binary classifiers constructed from the columns of the corresponding ECOC matrix whose rows consists of the class codewords \cite{db}. Each column defines a bipartition of the dataset by merging classes with the same bit value.  Decoding (classification) is performed by matching the codeword predicted by $L$ with the class codeword nearest in Hamming distance. In essence, ECOC is a generalization of one-vs-one and one-vs-all classification techniques, and as an ensemble technique, it is most effective when the binary classifiers make independent mistakes on a randon sample. 

 In this paper we derive new bounds on ECOC classification error rates that improve on that obtained by \cite{gs} by first applying the Feller and Chernoff bounds that are well known in statistics to the case where all binary classifiers are mutally independent and then applying a more recent bound due to \cite{kz} where they are correlated.  These new bounds theoretically establish the effectiveness of the ECOC approach in machine learning; in particular, we show under certain assumptions that ECOC classification error decays exponentially to zero with respect to codeword length.  We also present experimental results to demonstrate the validity of these bounds by applying them to various datasets to show the effect of correlation on classification accuracy.  
 
 Let $L=\{L_1,\ldots, L_n\}$ denote the aforementioned ensemble of $n$ binary classifiers (or learners) for a data set $S$ with $C$ classes.  Let $e_i$ denote the error rate of $L_i$.  Since $L_i$ is a binary classifier that only outputs 0 or 1 where $L_i=1$ indicating an error, we shall also call $e_i$ the bit error rate since the outputs $L_1,L_2,\ldots, L_n$ represent a binary string.  The following result, due to \cite{gs} gives a crude bound on the accuracy of $L$:
 
 \begin{theorem}[GS Bound, \cite{gs}] \label{th:gs_bound}
Let $\bar{e}=(e_1+\ldots+e_n)/n$ denote the average bit error rate.  Then the ECOC classification error rate $\mathcal{E}$ of $L$ is bounded by four times the average bit error rate, i.e.,
\begin{equation}\label{eq:main-gs-bound}
\mathcal{E} \leq \frac{4}{n}(e_1+\ldots+e_n) = 4
\bar{e}
\end{equation}
\end{theorem}
We note that the GS bound makes no assumptions regarding whether or not the classifiers are independent or how much correlation exists between them.  However, the GS bound is far from being sharp: assuming that $\bar{e}=0.1$, then $\mathcal{E}\leq 0.4$.  Thus, the GS bound fails to answer whether it is theoretically possible for $\mathcal{E} < \bar{e}$, which would validate its effectiveness as an ensemble technique.  Moreover, the GS bound gives no explicit dependence of $\mathcal{E}$ on $n$.

To the best of our knowledge and prior to this work, no error bound exists that rigorously demonstrates that $\mathcal{E} < \bar{e}$ is theoretically possible in the ECOC setting.  Progress so far has been limited to extending the GS bound to loss-based decoding schemes \cite{allwein} and special distance measures \cite{kjo}.  In addition, theorems have been proven that bound the excess error rate of the ECOC classifier in terms of the excess error rates of the constituent binary classifiers \cite{langford2005, beygelzimer2009}.  Here, ``excess errror rate" refers to the difference between the error rate and the Bayes optimal error rate.

Our main result establishes new bounds on $\mathcal{E}$ by calling on results  from statistical theory.

\begin{theorem}[Main Result] Let $M$ be the ECOC matrix corresponding to $L$ with row dimension $n$ and minimum row Hamming distance $2m$.
Set $r=m/n$ and $\bar{e}=\frac{1}{n}\sum_{i=1}^n e_i$ with $\bar{e} \neq r$.
\begin{enumerate}
\item
Chernoff Bound: If all binary classifiers are mutually independent, then
\begin{equation} \label{eq:main-chernoff-bound}
\mathcal{E} \leq \lambda^n
\end{equation}
where
$\displaystyle
\lambda=\frac{e^{r-\bar{e}}}{(r/\bar{e})^r}
$.
\item
KZ Bound: If $e_i=\bar{e}$ with $\displaystyle \bar{e} \leq \frac{m-1}{n-1}$, and all binary classifiers are mutually correlated up to second-order only and specified by a uniform non-negative correlation coefficient $c$ that satisfies the Bahadur bound (\ref{eq:bahadur}), then
\begin{equation} \label{eq:main-kz-bound}
\mathcal{E} \leq \lambda^n + 0.5 c n (n-1)\left(\frac{m-1}{n-1}-\bar{e}\right)\omega^n
\end{equation}
where $\lambda$ is defined in part 1 and
 $\displaystyle \omega=\left(\frac{\bar{e}}{r}\right)^r \left(\frac{1-\bar{e}}{1-r}\right)^{1-r}$.
\end{enumerate}
\end{theorem}

Assuming $r$ is fixed, these bounds imply that $\mathcal{E}$ decays exponentially to zero with respect to $n$ (codeword length).

\section{Indepdendent Base Classifiers}

In this section assume that all classifiers are mutually independent, but not necessarily identically distributed. This allows us to use the Poisson binomial distribution to describe the probability of error for our ensemble of classifiers and show that the corresponding ECOC error is bounded by the classical binomial distribution based on the maximum error rate of all the classifiers.  

Although the assumption of independence rarely holds in practice for real-world data sets, it is still useful as a starting point for our theoretical analysis and for establishing baseline results.  An important application where this assumption is considered involves the setting of multi-view learning within the context of co-training \cite{blum1998}, where say two
classifiers are trained separately on data representing two different views (or sets of attributes).  In this setting one of the assumptions requires the classifiers to be conditionally independent  given the class label.  This assumption can be relaxed \cite{dasgupta2001, balcan2004}.  We aim to do the same in the section where we take into account correlation between classifiers.

Denote by $S(n,k)$ the collection of all $k$-element subsets of $[n]=\{1,\ldots,n\}$.  Given a subset $A$ of $[n]$, we define the outcome $L_A$ to be such that $L_i=1$ if $i\in A$ and $L_i=0$ if $i\in \bar{A}$, where $\bar{A}$ denotes the complement of $A$ in $[n]$.  

\begin{definition}  Let $E=\{e_1,\ldots, e_n\}$ be a set of error rates of $\{L_1,\ldots, L_n\}$, respectively.  We define $p_E(n,k)$ to be the probability of the event where exactly $k$ out of the $n$ classifiers suffered bit errors, i.e., those outcomes $L_A$ where $\sum_{i=1}^n L_i = k$.  Then $p_E(n,k)$ is given by (Poisson binomial distribution)
\begin{equation} 
p_E(n,k)=\sum_{A\in S(n,k)} \left(\prod_{i\in A} e_i\right)\left(\prod_{j\in \bar{A}} (1-e_j)\right)
\end{equation}
If the classifiers are identically distributed so that $e_i=\bar{e}$ for all $i=1,\ldots, n$, then we define this probability by (binomial distribution)
\begin{equation}
p(n,k,\bar{e})= \binom{n}{k} \bar{e}^k (1-\bar{e})^{n-k}
\end{equation}
\end{definition}

Recall that the minimum Hamming distance between any two rows or any two columns of an $n$-dimensional Hadamard matrix $H$ is $n/2$ (see \cite{gs}).  In that case, when at least $n/4$ of the classifiers (corresponding to the columns of $H$) each makes an error, i.e., misclassifies a sample, then ECOC misclassification may occur.  This is because the rows of a $H$ describes an error-correcting code that only guarantees correct decoding up to (but strictly less than) $n/4$ bit errors.  Therefore, in order to bound $\mathcal{E}$, we shall assume under a worst-case scenario that misclassification always occur when $k\geq n/4$, where $k$ is the number of classifiers that suffered bit errors.

The following theorem shows that $p_E(n,k)$ can be bounded by the maximum error rate of all the classifiers, assuming all are no larger than $k/n$.

\begin{theorem} \label{th:prob}
Let $n,k\in \mathbb{N}$ with $0< k < n$.  Let $E=\{e_1,\ldots,e_n\}$ be a set of error rates with $0\leq e_i \leq k/n$ for all $i\in [n]$.  Set $e_{\max} = \max(E)$.  Then
\begin{equation} \label{eq:error-bound}
p_E(n,k) \leq p(n,k,e_{\max})
\end{equation}

\end{theorem}

The proof of this theorem requires the following lemmas, whose proofs are given in the appendix of this paper \cite{nguyen2021appendix}.  Before stating them, we first introduce notation: given $m \in [n]$, we define $A_m=[n]-\{m\}$ and $E_m=E-\{e_m\}$.   

\begin{lemma} \label{le:inequality}
We have
\begin{equation}  \label{eq:inequality}
p_{E_m}(n-1,k-1) - p_{E_m}(n-1,k)  > 0
 \end{equation}
 for all $m \in [n]$ and $k=2,\ldots, n$.
\end{lemma}

\begin{lemma} 
\label{le:monotone}
$p_E(n,k)$ is strictly increasing with respect to $e_i$ over the interval $(0,k/n)$.

\end{lemma}

\begin{proof}(of Theorem \ref{th:prob}) Since $p_E(n,k)$ is monotone increasing in each variable $e_m$, it is maximal when each $e_m$ is replaced by $e_{\max}$.  Thus,
\[
p_E(n,k) \leq p(n,k,e_{\max})
\]
as desired.
\end{proof}

\begin{definition}  We define the \textit{maximum} ECOC error rate $\varepsilon_E(n,m)$ as the probability of the event where at least $m$ out of $n$ independent binary classifiers produces an error and is given by the cumulative sum
\begin{equation} 
\varepsilon_E(n,m)=\sum_{k=m} ^n p_E(n,k)
\end{equation}
If the classifiers are identically distributed, then we define the probability of this event by
\begin{equation}
\label{eq:error-independent}
\varepsilon(n,m,\bar{e})=\sum_{k=m}^n p(n,k,\bar{e})
\end{equation}
\end{definition}

It is clear that $\mathcal{E} \leq \epsilon_E(n,m)$.  Moreover, we note that $\varepsilon_E(n, n/4)$ gives the maximum ECOC error rate for a Hadamard matrix $M$ of dimension $n=4m$ with minimum row Hamming distance $n/2$.  The following theorem, which follows immediately from Theorem \ref{th:prob}, shows that $\varepsilon_E(n,m)$ is bounded by the binomial distribution based on the largest bit error rate.

\begin{theorem} \label{th:bound_ecoc_error}
Suppose $0\leq e_i \leq m/n$ for all $i\in [n]$.    Set $e_{\max} = \max(E)$.  Then
\begin{equation} \label{eq:binomial-bound}
\varepsilon_E(n,m) \leq \varepsilon(n,m ,e_{\max})
\end{equation}
\end{theorem}

We now apply Feller's result on $\varepsilon(n,m,\bar{e})$ to obtain the following simple rational bound:

\begin{lemma}[\cite{feller}]  \label{le:bound}
For $m > n\bar{e}$, we have
\begin{equation} \label{eq:simple-bound}
\varepsilon(n,m,\bar{e})\leq \frac{m(1-\bar{e})}{(m-n\bar{e})^2}
\end{equation}
\end{lemma}

The following corollary shows that ECOC error rate tends to zero as the codeword length tends to infinity assuming the ratio $m/n$ stays fixed.  This gives theoretical justification for the effectiveness of the ECOC approach for datasets with a large number of classes; of course, this assumes the existence of many relatively accurate independent binary classifiers.

\begin{corollary} Suppose $r=m/n$ and $\hat{e}$ are fixed with $0\leq e_i \leq \hat{e} < r$ for all $i\in [n]$.  Then
\begin{equation}
\lim_{n\rightarrow \infty} \mathcal{E}=0
\end{equation}
\end{corollary}

\begin{proof}  Set $e_{\max}=\max(E)$. Then $e_{\max}\leq \hat{e}$ and since $\hat{e}<r$, we have $m>n\hat{e}$. It follows from Theorem \ref{th:bound_ecoc_error} and Lemma \ref{le:bound} that the chain of inequalities hold:
\begin{align*}
\mathcal{E} & \leq  \varepsilon_E(n,m) \leq \varepsilon(n,m,e_{\max}) \leq \varepsilon(n,m,\hat{e}) \\ & \leq \frac{m(1-\hat{e})}{(m-n\hat{e})^2} \leq  \frac{1}{n}\left[\frac{r(1-\hat{e})}{(r-\hat{e})^2}\right]
\end{align*}
It is now clear that $\mathcal{E} \rightarrow 0$ as $n\rightarrow \infty$.
\end{proof}

To obtain a sharper and more useful bound, we call on the following result by Chernoff.

\begin{theorem}[\cite{chernoff}] Let $\mu=\sum_{i=1}^n e_i$.  Then
\begin{equation} \label{eq:chernoff-bound}
\varepsilon_E(n,m) \leq \frac{e^{m-\mu}}{(m/\mu)^m}
\end{equation}
where $e$ is the Euler number.
\end{theorem}

The following corollary, which restates the Chernoff bound in terms of the average bit error rate, shows that $\mathcal{E}$ decays to zero exponentially with respect to codeword length.
\begin{corollary} \label{co:new_bound}
Let $r=m/n$, $\bar{e}=\frac{1}{n}\sum_{i=1}^n e_i$, and
$\displaystyle
\lambda=\frac{e^{r-\bar{e}}}{(r/\bar{e})^r}
$.
Then
\begin{equation}
\label{eq:lambda-bound}
\varepsilon_E(n,m) \leq \lambda^n
\end{equation}
where $0\leq \lambda < 1$ for all $\bar{e}\neq r$.  Moreover, $\lambda$ is increasing with respect to $\bar{e}$ for $0\leq \bar{e}<r$ and decreasing with respect to $r$ for $r < \bar{e} \leq 1$.  Thus, if $r$ and $\bar{e}$ are fixed with $\bar{e}\neq r$, then $\varepsilon_E(m,n)$, and thus $\mathcal{E}$, decays exponentially to zero as $n\rightarrow \infty$.
\end{corollary}

\begin{proof}
It is straightforward to prove using analytical methods that $0\leq \lambda \leq 1$ and that $\lambda$ is increasing and decreasing with respect to $\bar{e}$ over the respective intervals.  As for the bound (\ref{eq:lambda-bound}), we have
\begin{align}
\varepsilon_E(n,m) \leq \frac{e^{m-\mu}}{(m/\mu)^m} \leq \frac{e^\frac{{r-\bar{e}}}{n}}{(r/\bar{e})^{rn}} = \left(\frac{e^{r-\bar{e}}}{(r/\bar{e})^r}\right)^n =\lambda^n
\end{align}
Since $\bar{e}\neq r$ due to $\lambda <1$, it follows that $\epsilon_E(n,m)$ (and thus $\mathcal{E}$) decays exponentially as $n\rightarrow \infty$.
\end{proof}

\begin{figure}
\centering
\includegraphics[width=250pt]{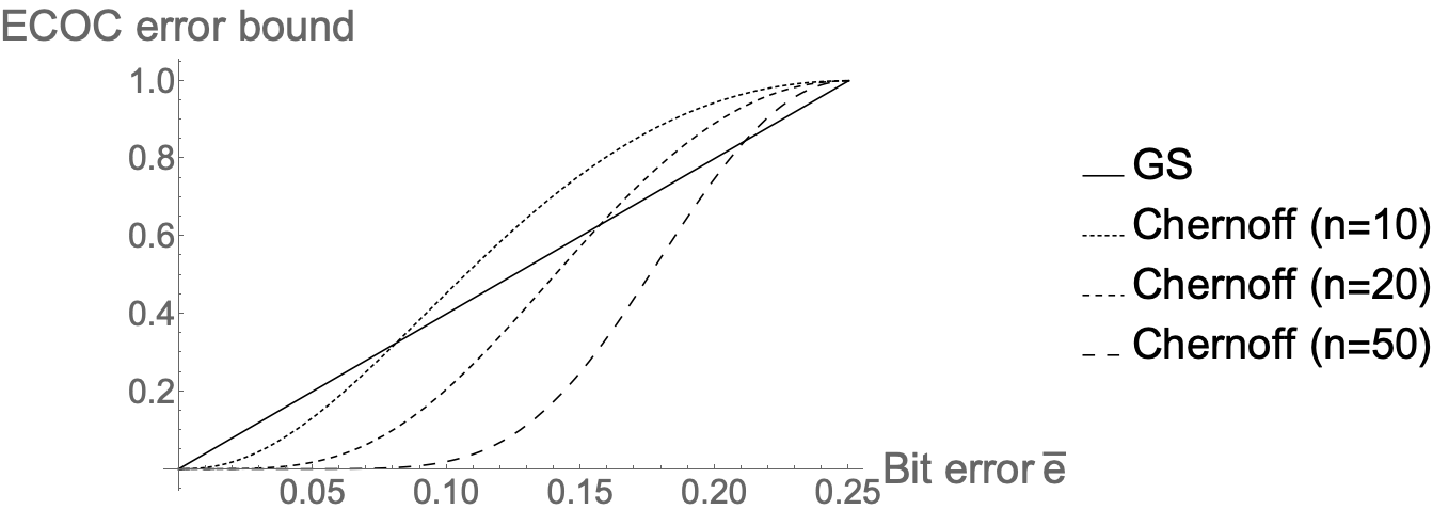}
\caption{ECOC error bounds: GS vs Chernoff ($n=10,20,50$)}
\label{fig:bounds-gs-chernoff}
\end{figure}

We emphasize that the improvement of the Chernoff bound (Corollary \ref{co:new_bound})  over the GS bound (Theorem \ref{th:gs_bound}) is due to the assumption that all the binary classifiers are mutually independent.   Figure \ref{fig:bounds-gs-chernoff} clearly demonstrates this for large $n$ and small $\bar{e}$.  

We end this section by commenting that Corollary \ref{co:new_bound} is also valid for the non-binary ECOC setting where the coefficients of the ECOC matrix is chosen from a non-binary alphaset \cite{zhou2019}. 

\section{Correlated Base Classifiers}

In this section we assume dependence (correlation) between certain base classifiers to show how it affects ECOC accuracy.  We first make the simple assumption that all binary classifiers $\{L_1,\ldots, L_n\}$ are mutually independent except for a pair of dependent classifiers $L_{n-1}$ and $L_n$, which are allowed to depend on each other as follows.  Recall that each $L_i$ takes on two possible values, namely $L_i= 0$ (correct prediction) and $L_i=1$ (incorrect prediction).  As before, let $e_i$ denote the error rate of $L_i$, i.e., $P(L_i=1)=e_i$.  Since $L_{n-1}$ and $L_n$ are dependent on each other, we specify their correlation via the joint probability
\begin{equation}
P_{11}:=P(L_{n-1}=1 \ \textrm{and} \ L_n=1)=f
\end{equation}
It follows that the remaining joint probabilities are given by
\begin{align}
P_{10}:=P(L_{n-1}=0 \ \textrm{and} \ L_n=1) & =e_{n-1}-f \\
P_{01}:=P(L_{n-1}=1 \ \textrm{and} \ L_n=0) & =e_n-f \\
P_{00}:=P(L_{n-1}=1 \ \textrm{and} \ L_n=1) & =1-e_{n-1}-e_n+f
\end{align}
We shall assume that $0\leq e_{n-1}\leq 1/2$, $0 \leq e_n \leq 1/2$, and $0 \leq f\leq \min(e_{n-1},e_n)$ so that all probabilities are non-negative.  We then define the correlation between $L_{n-1}$ and $L_n$ as 
\begin{equation}
c=\textrm{cor}(L_{n-1},L_n) = \frac{f-e_{n-1}e_n}{\sqrt{e_{n-1}(1-e_{n-1})e_n(1-e_n)}}
\end{equation}
In particular, if $L_{n-1}$ and $L_n$ are independent so that $f=e_{n-1}e_n$, then $c=0$.

Given a subset $A\in S(n,k)$, we denote $\bar{A}=[n]-A$ and define
\begin{equation}
P_n(A):=P(\{L_i=1:i\in A\} \ \textrm{and} \  \{L_i=0:i\in \bar{A}\})
\end{equation}
\begin{definition} Let $E_n=\{e_1,\ldots, e_n\}$.  We define the probability of the event where $k$ out of $n$ classifiers produces an error (with dependence between classifiers $L_{n-1}$ and $L_n$ as defined above) by
\begin{equation}
p_{E_n}(n,k,f)=\sum_{A\in S(n,k)} P(A).
\end{equation}
If $e_i=\bar{e}$ for all $i\in [n]$, then we denote $p(n,k,\bar{e},f):=p_{E_n}(n,k,f)$.

\end{definition}

Define $\bar{e}_n=(e_{n-1}+e_n)/2$.  The following lemma, whose proof is given in 
\cite{nguyen2021appendix}, shows the explicit dependence of $p_{E_n}(n,k,f)$ on $\bar{e}_n$ and $f$.
\begin{lemma} \label{le:pcor} We have
\begin{equation} \label{eq:pcor}
p_{E_n}(n,k,f)
=
\begin{cases}
A, & \textrm{if} \ k \leq n-2; \\
B, & \textrm{if} \ k = n-1; \\
C, & \textrm{if} \ k = n.
\end{cases}
\end{equation}
where
\begin{align*}
A & =
 fp_{E_{n-2}}(n-2,k-2) +2(\bar{e}_n-f)p_{E_{n-2}}(n-2,k-1)  \\
& \ \ \ \ + (1-2\bar{e}_n+f)p_{E_{n-2}}(n-2,k)  \\
B & = fp_{E_{n-2}}(n-2,n-3) +2(\bar{e}_n-f)p_{E_{n-2}}(n-2,n-2) \\
C &= fp_{E_{n-2}}(n-2,n-2) 
\end{align*}
\end{lemma}

Define $\hat{e}_{\max}=\max(E_{n-2})$.  We apply Theorem \ref{th:prob} to the above lemma to obtain the following bound.

\begin{corollary}
Suppose $\hat{e}_{\max} \leq \frac{k-2}{n-2}$.  Then
\begin{equation}
p_{E_n}(n,k,f) \leq
\begin{cases}
A, & \textrm{if} \ k \leq n-2; \\
B, & \textrm{if} \ k = n-1; \\
C, & \textrm{if} \ k = n.
\end{cases}
\end{equation}
where
\begin{align*}
A &=
fp(n-2,k-2,\hat{e}_{\max})\\
& \ \ \ \ +2(\bar{e}_n-f)p(n-2,k-1,\hat{e}_{\max}) & \\
& \ \ \ \ + (1-2\bar{e}_n+f)p(n-2,k,\hat{e}_{\max}) \\
B &=fp(n-2,n-3,\hat{e}_{\max}) \\
& \ \ \ \ +2(\bar{e}_n-f)p(n-2,n-2,\hat{e}_{\max}) \\
C &=fp(n-2,n-2,\hat{e}_{\max})
\end{align*}
In the special case where all binary classifiers are identically distributed, i.e., $e_i=\bar{e}$ for all $i\in [n]$, then
\begin{align*}
A & = fp(n-2,k-2,\bar{e})+2(\bar{e}-f)p(n-2,k-1,\bar{e}) & \\
& \ \ \ \ + (1-2\bar{e}+f)p(n-2,k,\bar{e})  \\
B & = fp(n-2,n-3,\bar{e})+2(\bar{e}-f)p(n-2,n-2,\bar{e})  \\
C & = fp(n-2,n-2,\bar{e})
\end{align*}
\end{corollary}

The next lemma, whose proof is given in \cite{nguyen2021appendix}, assumes all classifiers are identically distributed.

\begin{lemma} \label{le:pincreasing}
$p(n,k,\bar{e},f)$ is increasing with respect to $f$ for fixed $\bar{e}\in I$, where
\[
I=
\begin{cases}
[0,k/n-\alpha(n,k)] & \textrm{if} \ k \leq n-2; \\
[0,1-2/n] & \textrm{if} \ k = n-1; \\
[0,1] & \textrm{if} \ k = n,
\end{cases}
\]
and $\displaystyle \alpha(n,k)=\frac{1}{n}\sqrt{\frac{k(n-k)}{n-1}}$.
\end{lemma}

\begin{definition}  We define the maximum ECOC error rate $\varepsilon_E(n,m,f)$ (assuming correlation given by $f$) as the probability of the event where at least $m$ out of $n$ binary classifiers produces an error and therefore is given by the cumulative sum
\begin{equation} 
\varepsilon_E(n,m,f)=\sum_{k=m} ^n p_E(n,k,f)
\end{equation}
If the classifiers are identically distributed, i.e., $e_i=\bar{e}$ for all $i=1,\ldots, n$, then we define
\begin{equation}
\varepsilon(n,m,\bar{e},f):=\varepsilon_E(n,m,f)=\sum_{k=m}^n p(n,k,\bar{e},f)
\end{equation}
\end{definition}

The next two theorems describe the dependence of the maximum ECOC error rate on $f$.  Their proofs can be found in \cite{nguyen2021appendix}.

\begin{theorem} 
\label{th:epsilon-correlation}
We have
\begin{align}
\varepsilon(n,m,\bar{e},f) & =f\varepsilon(n-2,m-2,\bar{e}) \nonumber \\
& \ \ \ \ +2(\bar{e}-f)\varepsilon(n-2,m-1,\bar{e}) \nonumber \\
& \ \ \ \ + (1-2\bar{e} +f)\varepsilon(n-2,m,\bar{e})
\end{align}

\end{theorem}

\begin{theorem}
\label{th:correlation-monotone}
$\varepsilon(n,m,\bar{e},f)$ is increasing with respect to $f$ for $\displaystyle 0\leq \bar{e} \leq \frac{m-1}{n-1}$ and decreasing with respect to $f$ for $\displaystyle \frac{m-1}{n-1} \leq \bar{e} \leq 1$.

\end{theorem}

The next theorem gives a simple bound for $\varepsilon(n,m,\bar{e},f)$, which again implies that $\mathcal{E}$ decays exponentially to zero but assumes that $\bar{e}$ is fixed.

\begin{theorem}
Let $r=\frac{m-2}{n-2}$ and $\lambda=\frac{e^{r-\bar{e}}}{(r/\bar{e})^r}$.  Then
\begin{equation}
\varepsilon(n,m,\bar{e},f) \leq \lambda^{n-2}
\end{equation}
\end{theorem}

\begin{proof} We apply Theorem (\ref{th:epsilon-correlation}) and Corollary \ref{co:new_bound}:
\begin{align} 
\varepsilon(n,m,\bar{e},f) &  \leq  f\cdot \lambda_1^{n-2} + 2(\bar{e}-f)\cdot \lambda_2^{n-2} \nonumber \\ 
& \ \ \ \ + (1-2\bar{e}+f)\cdot \lambda_3^{n-2} \\
& \leq \lambda_1^{n-2}
\end{align}
since $\lambda_1\geq \lambda_2 \geq \lambda_3$.  Setting $\lambda=\lambda_1$ gives the desired result.
\end{proof}

Let us now use Theorem \ref{th:correlation-monotone} to discuss the effect of a correlated pair of binary classifiers on the maximum ECOC error rate  $\varepsilon(n,m,e,f)$.  Assuming $0\leq \bar{e} \leq \frac{m-1}{n-1}$, which implies $\varepsilon(n,m,\bar{e},f)$ is increasing with respect to $f$, we conclude that over the range $0\leq f < \bar{e}$, the ECOC error rate is lower when there is negative correlation ($f < \bar{e}^2$) compared to that for independence ($f=\bar{e}^2$), which in turn is lower than when there is positive correlation ($f> \bar{e}^2$).   In other words, having negative correlation actually helps to decrease the ECOC accuracy while having positive correlation increases the ECOC error rate, which agrees with our common intuition.  On the other hand, over the range $\frac{m-1}{n-1} \leq \bar{e} \leq 1$, the reverse occurs since $\varepsilon(n,m,\bar{e},f)$ is decreasing with respect to $f$.  Thus, the moral is that positive correlation is detrimnental only if $\bar{e}$ is relatively small .

We next investigate the effect of having all classifiers mutually dependent on ECOC accuracy.

\subsection{All Classifiers Mutally Correlated}

Suppose all classifiers are mutally correlated up to second-order only (all higher-order correlations are zero).  We define

\begin{align}
Z_i & = \frac{L_i - e_i}{\sqrt{e_i(1 - e_i)}} \\
f_{ij} & = P(L_i=1 \ \textrm{and} \ L_j=1) \\
c_{ij} & = \mathrm{cor}(Z_i, Z_j) =\frac{f_{ij}-e_i e_j}{\sqrt{e_i(1-e_i)e_j(1-e_j)}} \label{eq:correlation}
\end{align}

Let $A=\{a_1,\ldots, a_k\} \in S(n,k)$.  Recall our definition of the outcome $L_A$ where $L_i=1$ if $i\in A$ and $L_i=0$ if $i\in \bar{A}$ where $k=\sum_{i=1}^n L_i$.   Denote by $\pi_A$ the probability of the outcome $L_A$.

\begin{definition} Suppose $e_i=\bar{e}$ and $c_{ij}=c$.   We define the probability of the event where $k$ out of $n$ classifiers produces an error (with correlation given by (\ref{eq:correlation})) by
\begin{equation}
p(n,k,\bar{e},c)=\sum_{A\in S(n,k)} \pi_A.
\end{equation}
We also define the maximum ECOC error rate $\varepsilon(n,m,\bar{e},c)$ as the probability of the event where at least $m$ out of $n$ binary classifiers produces an error and therefore is given by the cumulative sum
\begin{equation} 
\varepsilon(n,m,\bar{e},c)=\sum_{k=m} ^n p(n,k,\bar{e},c)
\end{equation}
\end{definition}

The following result by \cite{kz} gives an explicit formula for $\varepsilon(n,m)$ where $e_i=\bar{e}$ are equal and all correlations $c_{ij}=c$ are equal.  Although their result is stated under the assumption $0.5 \leq \bar{e} \leq 1$ because of its application to jury design where in their model jurors are assumed to be competent, their proof, which we partially replicate in \cite{nguyen2021appendix} for completeness, in fact holds over the range $0\leq \bar{e} \leq 1$, assuming that $c$ satisfies the Bahadur bound described in the same paper:
\begin{equation}\label{eq:bahadur}
-\frac{2(1-\bar{e})}{n(n-1)\bar{e}} \leq c \leq 
\frac{2\bar{e}(1-\bar{e})}{(n-1)\bar{e}(1-\bar{e})+0.25-\gamma},
\end{equation}
where
\begin{equation}
\gamma =\min_{0\leq k \leq n}\{[k-(n-1)\bar{e}-0.5]^2\}\leq 0.25
\end{equation}

\begin{theorem}[\cite{kz}]
\label{th:kz}
Suppose $e_i=\bar{e}$ for all $1\leq i \leq n$ and $c_{ij}=c$ for all $1\leq i,j \leq n$, and that $c$ satisfies (\ref{eq:bahadur}).  Then
\begin{align}
\label{eq:kz}
\varepsilon(n,m,\bar{e},c) =\varepsilon(n,m,\bar{e}) \hspace{110pt} \nonumber \\
 +
0.5 c n (n-1)\left(\frac{m-1}{n-1}-\bar{e}\right)p(n-1,m-1,\bar{e})
\end{align}
where $\varepsilon(n,m,\bar{e})$ is defined by (\ref{eq:error-independent}) and
\begin{equation}
p(n-1,m-1,\bar{e})=\binom{n-1}{m-1}\bar{e}^{m-1}(1-\bar{e})^{n-m}
\end{equation}
\end{theorem}

\begin{corollary} Let $r=m/n$.  Suppose $\displaystyle \bar{e} \leq \frac{m-1}{n-1}$ and $c$ is non-negative and satisfies (\ref{eq:bahadur}). Then
\begin{equation}
\varepsilon(n,m,\bar{e},c) \leq \lambda^n + 0.5 c n (n-1)\left(\frac{m-1}{n-1}-\bar{e}\right)\omega^n
\end{equation}
where
$\displaystyle \lambda=\frac{e^{r-\bar{e}}}{(r/\bar{e})^r}$ and $\displaystyle \omega=\left(\frac{\bar{e}}{r}\right)^r \left(\frac{1-\bar{e}}{1-r}\right)^{1-r}$.
Moreover, if $r$ and $\bar{e}$ are fixed, then $\varepsilon(n,m,\bar{e},c)$ (and thus $\mathcal{E}$) decays exponentially to zero as $n \rightarrow \infty$.

\end{corollary}

\begin{proof} Since it was proven earlier that the first term on the right-hand side of (\ref{eq:kz}), $\varepsilon(n,m,\bar{e})$, is bounded by $\lambda^n$ and decays exponentially to zero as $n \rightarrow \infty$, it suffices to prove that the second term, $p(n-1,m-1,\bar{e})$, is bounded similarly.  We first manipulate it as follows:
\begin{align*}
p(n-1,m-1,\bar{e}) & =\binom{n-1}{m-1}\bar{e}^{m-1}(1-\bar{e})^{n-m} \\
& = \frac{(n-1)!}{(m-1)!(n-m)!}\frac{\bar{e}^m(1-\bar{e})^{n-m}}{\bar{e}} \\
& = \frac{m}{n} \binom{n}{m} \frac{\bar{e}^m(1-\bar{e})^{n-m}}{\bar{e}} \\
& = \frac{r}{\bar{e}} \binom{n}{m} \bar{e}^m(1-\bar{e})^{n-m}
\end{align*}
Then using the bound
\begin{equation}
\binom{n}{m} \leq \left( \left(\frac{m}{n}\right)^m\left(1-\frac{m}{n}\right)^{n-m}\right)^{-1},
\end{equation} we have
\begin{align*}
p(n-1,m-1,\bar{e}) & \leq \left(\frac{r}{\bar{e}}\right) \frac{\bar{e}^m(1-\bar{e})^{n-m} }{(m/n)^m(1-m/n)^{n-m} } \\
& \leq \left(\frac{r}{\bar{e}}\right) \left(\frac{\bar{e}}{r}\right)^m\left(\frac{1-\bar{e}}{1-r}\right)^{n-m} \\
& \leq \left(\frac{r}{\bar{e}}\right) \left( \left(\frac{\bar{e}}{r}\right)^r \left(\frac{1-\bar{e}}{1-r}\right)^{1-r}\right)^n
\end{align*}
Since $ \left(\frac{\bar{e}}{r}\right)^r \left(\frac{1-\bar{e}}{1-r}\right)^{1-r} < 1$ for $\bar{e} \neq r$, it follows that $p(n-1,m-1,\bar{e}) \rightarrow 0$ exponentially as $n \rightarrow \infty$.  Thus, the same holds for $\varepsilon(n,m,\bar{e},c)$ as well.
\end{proof}

\section{Experimental Results}

In this section we present experimental results to demonstrate the validity of our work by performing ECOC classification on various data sets and comparing the resulting classification error rates with those predicted by the Chernoff and KZ bounds established in the previous section.

In particular, we selected six public datasets to perform ECOC classification: Pendigits, Usps, Vowel, Letter Recognition (Letters), CIFAR-10, and Street View House Numbers (SVHN).  Information regarding these datasets are given in Table \ref{tab:dataset}. 

\begin{table}[htbp]
\begin{center}
\begin{tabular}{|c|c|c|c|c|c|} \hline
Dataset & \# Samples & \# Features & \# Classes ($C$) & $r=m/n$ \\
\hline 
Pendigits & 3498  & 16 & 10 & 2/11 \\
Usps & 7291 & 256 & 10 & 2/10 \\
Vowel & 990 & 10 & 11 & 2/11 \\
Letters & 20,000 & 16 & 26 & 6/26 \\
CIFAR-10 & 60,000 & Image & 10 & 2/10 \\
SVHN & 99,289 & Image & 10 & 2/10 \\
        \hline
\end{tabular}
\caption{\label{tab:dataset} Datasets}
\end{center}
\end{table}

\begin{enumerate}
    \item ECOC Matrix: We employed a square ECOC matrix $M$ for every dataset ($n=C$) and constructed $M$ from a $\{0,1\}$-Hadamard matrix $H$ of dimension $2^k$, where $k$ was chosen to be the smallest integer for which $2^k\geq c$ and $c$ denotes the number of classes.  We then truncated an appropriate number of rows and columns from $H$ (starting from the top left) to obtain our square matrix $M$ of dimension $n\times n$.  The parameter $r=m/n$ for each dataset is given in Table \ref{tab:dataset}.

\item Classification algorithms: For the datasets Pendigits, Usps, Vowel, and Letters, we employed two different models $L$ for our base classifiers: decision tree (DT) and support-vector machines (SVM), using the Python modules (version 3.7) sklearn.tree.DecisionTreeClassifier and sklearn.svm.SVC with default settings, respectively, utilizing the scikit-learn machine learning library.  Computations were performed on a standard laptop.  For the image datasets CIFAR-10 and SVHN, we employed a pre-trained convolutional neural network, Resnet-18 (loaded from Pytorch), with an additional dense layer to product binary output and using the Adam optimizer.  Computations were performed for 10 epochs with a batch size of 128 and ran on the Open Science Grid \cite{osg07,osg09}. 

\end{enumerate}

Thus, given a dataset, we performed 10th-fold cross-validation based on the experimental setup described above and recorded the ECOC error rate (experimental) for each fold, as well mean ECOC error rate and standard deviation for all ten folds.  To compute the GS, Chernoff, and KZ bounds given by (\ref{eq:main-gs-bound}), (\ref{eq:main-chernoff-bound}), and (\ref{eq:main-kz-bound}), respectively, for each fold, we used the mean bit error rate  $\bar{e}$, obtained by averaging the bit error rates of all the binary classifiers.  In addition, for the KZ bound we used the mean correlation for $c$, obtained by averaging the coefficients of the correlation matrix of the binary classifiers.  Full results, including values used for $\bar{e}$ and $c$, are given in \cite{nguyen2021appendix} (Tables III-VIII).

\subsection{Results and Discussion}

Experimental results show that for all datasets the ECOC error rates ($\mathcal{E}$) are either below all three bounds (GS, Chernoff, and KZ) or clustered around the Chernoff and KZ bounds, where the latter occurs for Letters (DT and SVM) and Pendigits (SVM).  This can be seen in the plots in Figures 2-7 for Pendigits, Letters, CIFAR-10, and SVHN (see \cite{nguyen2021appendix} for plots of USPS and Vowels) where ECOC error rates are shown for each of the ten folds and in Table \ref{tab:error-rate} where results are averaged over all folds (lowest value indicated in bold).  These results demonstrate the validity of all three bounds.  However, Figures 3-5 (Letters) and 7 (Pendigits) clearly show that the Chernoff and KZ bounds provide much more accurate estimates of the ECOC error compared to the GS bound.  This is to be expected for Letters where the number of binary classifiers ($n=26$), is signficantly larger than all the other datasets.  As discussed earlier, the Chernoff and KZ bounds decay exponentially to zero with respect to $n$ and thus are more effective for larger values of $n$. Overall, we believe our experimental results demonstrate that the Chernoff and GS bounds are quite useful in practice.


\begin{figure}
\includegraphics[width=240pt, height=165pt]{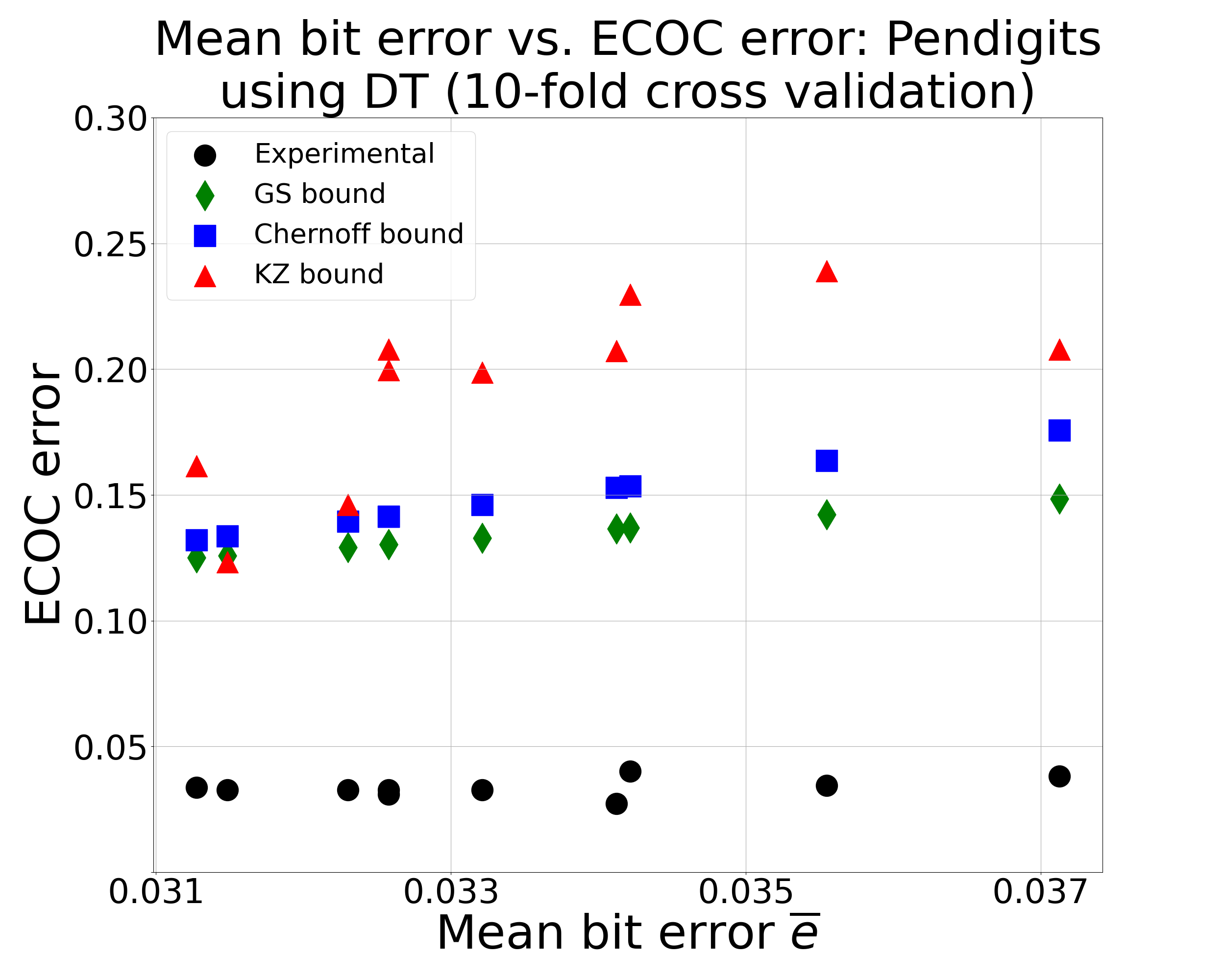}
\caption{Pendigits: Mean bit error vs ECOC \\ error (DT using 10-fold cross-validation)}
\label{fig:digits-dt}
\end{figure}

\begin{figure}
\includegraphics[width=240pt, height=165pt]{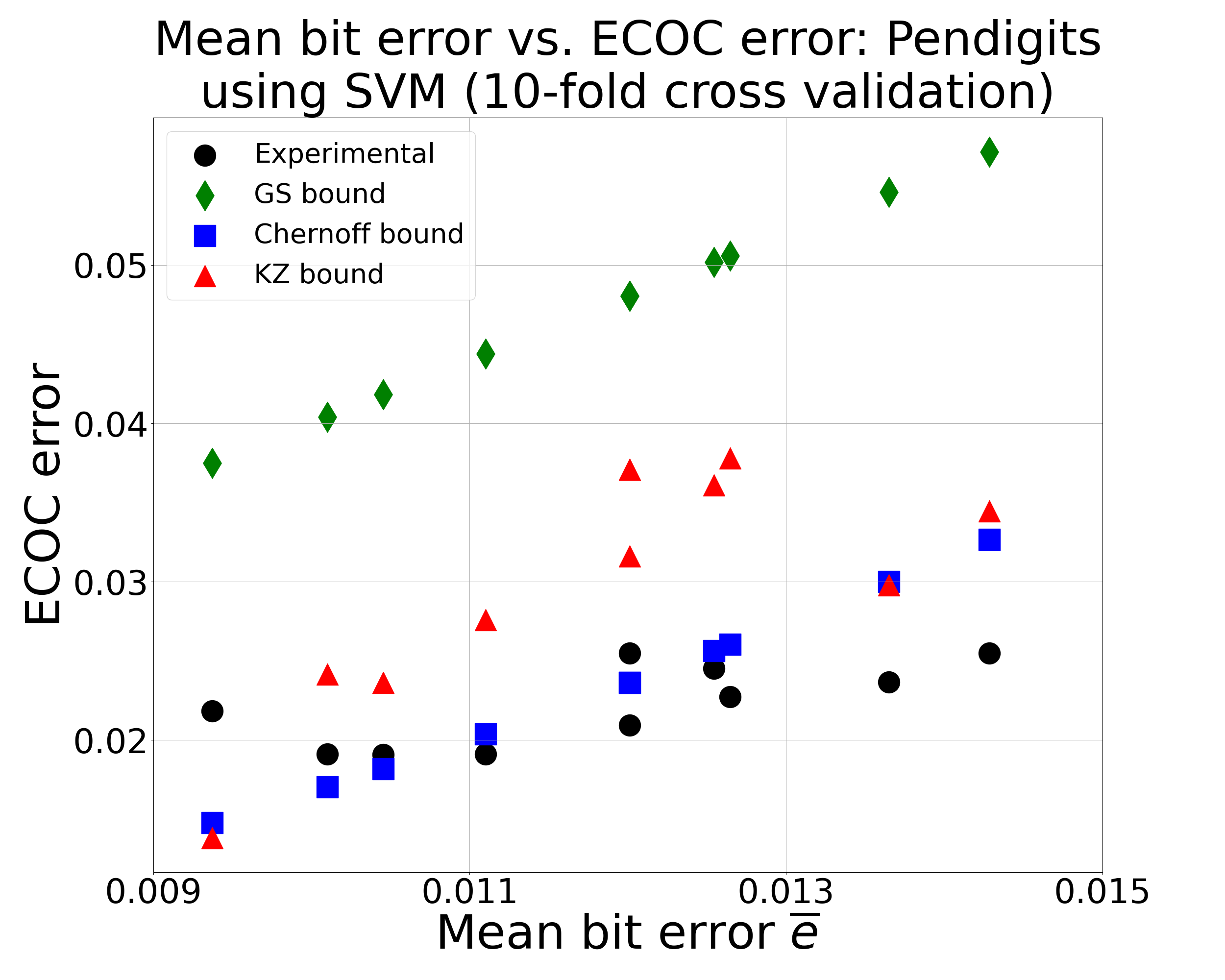}
\caption{Pendigits: Mean bit error vs ECOC \\ error (SVM using 10-fold cross-validation)}
\label{fig:digits-svm}
\end{figure}

\begin{figure}
\includegraphics[width=240pt, height=165pt]{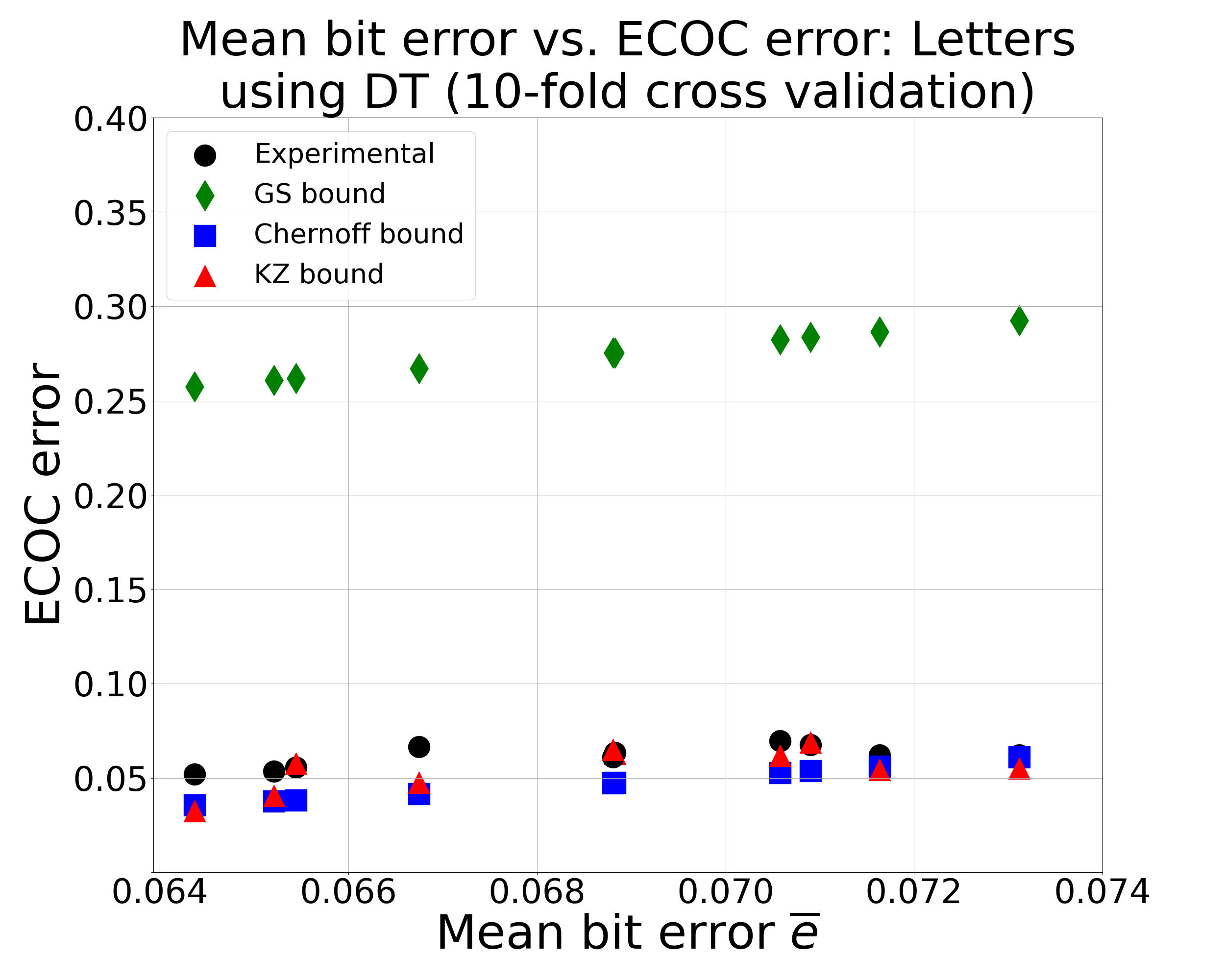}
\caption{Letters: Mean bit error vs ECOC \\ error (DT using 10-fold cross-validation)}
\label{fig:letters-dt}
\end{figure}

\begin{figure}
\includegraphics[width=240pt, height=165pt]{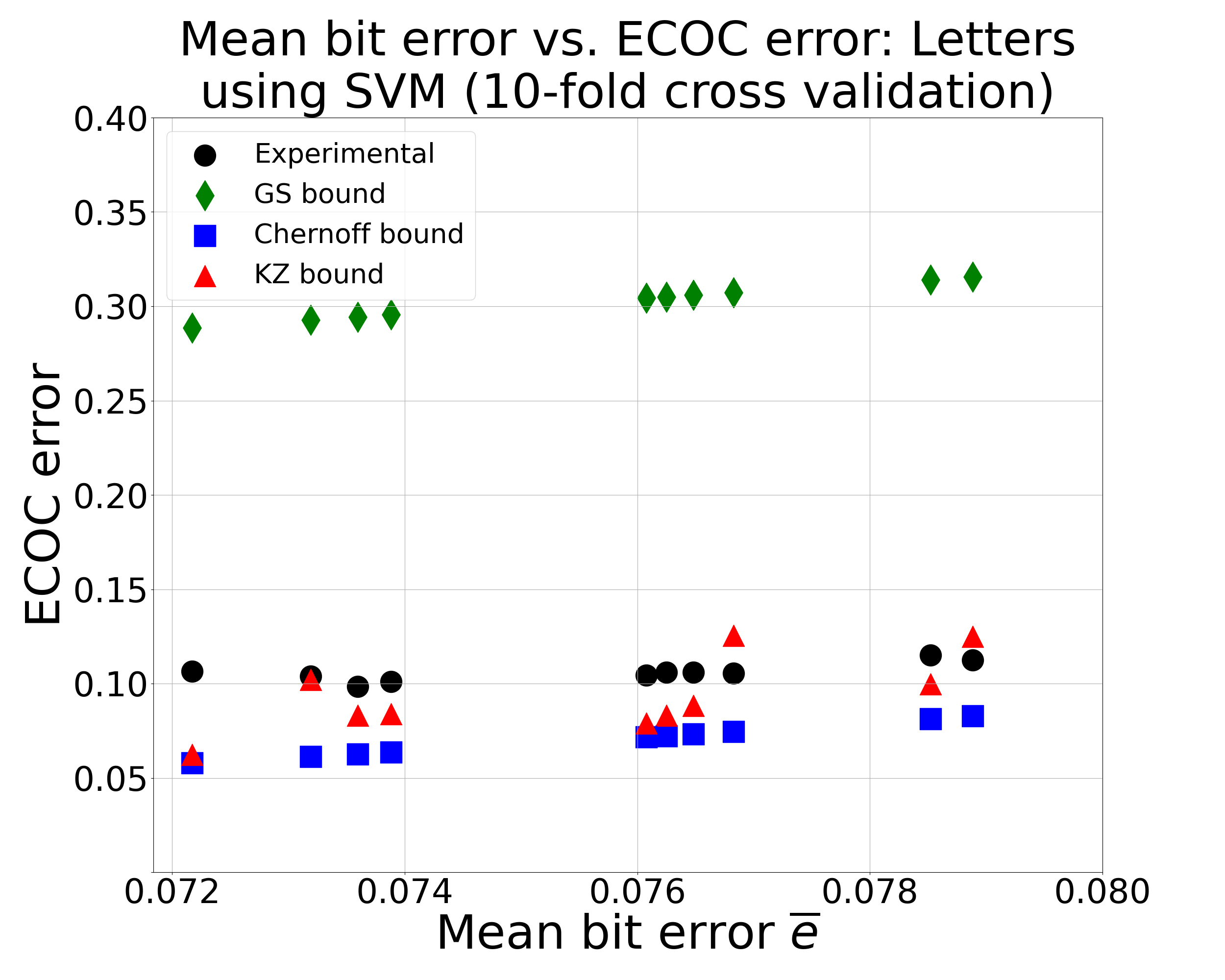}
\caption{Letters: Mean bit error vs ECOC \\ error (SVM using 10-fold cross-validation)}
\label{fig:letters-svm}
\end{figure}

\begin{figure}
\includegraphics[width=240pt, height=165pt]{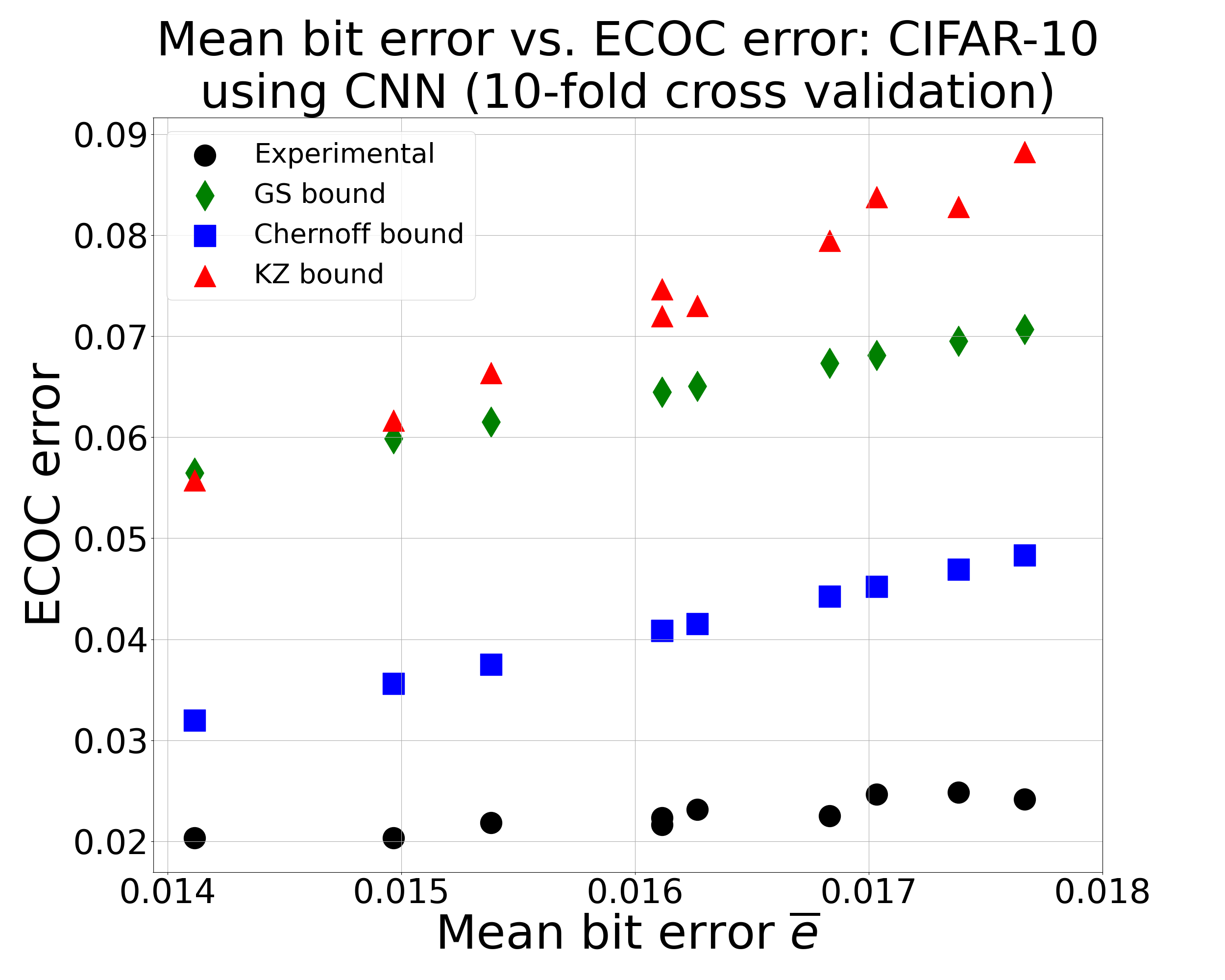}
\caption{CIFAR-10: Mean bit error vs ECOC \\ error (CNN using 10-fold cross-validation)}
\label{fig:cifar-dt}
\end{figure}

\begin{figure}
\includegraphics[width=240pt, height=165pt]{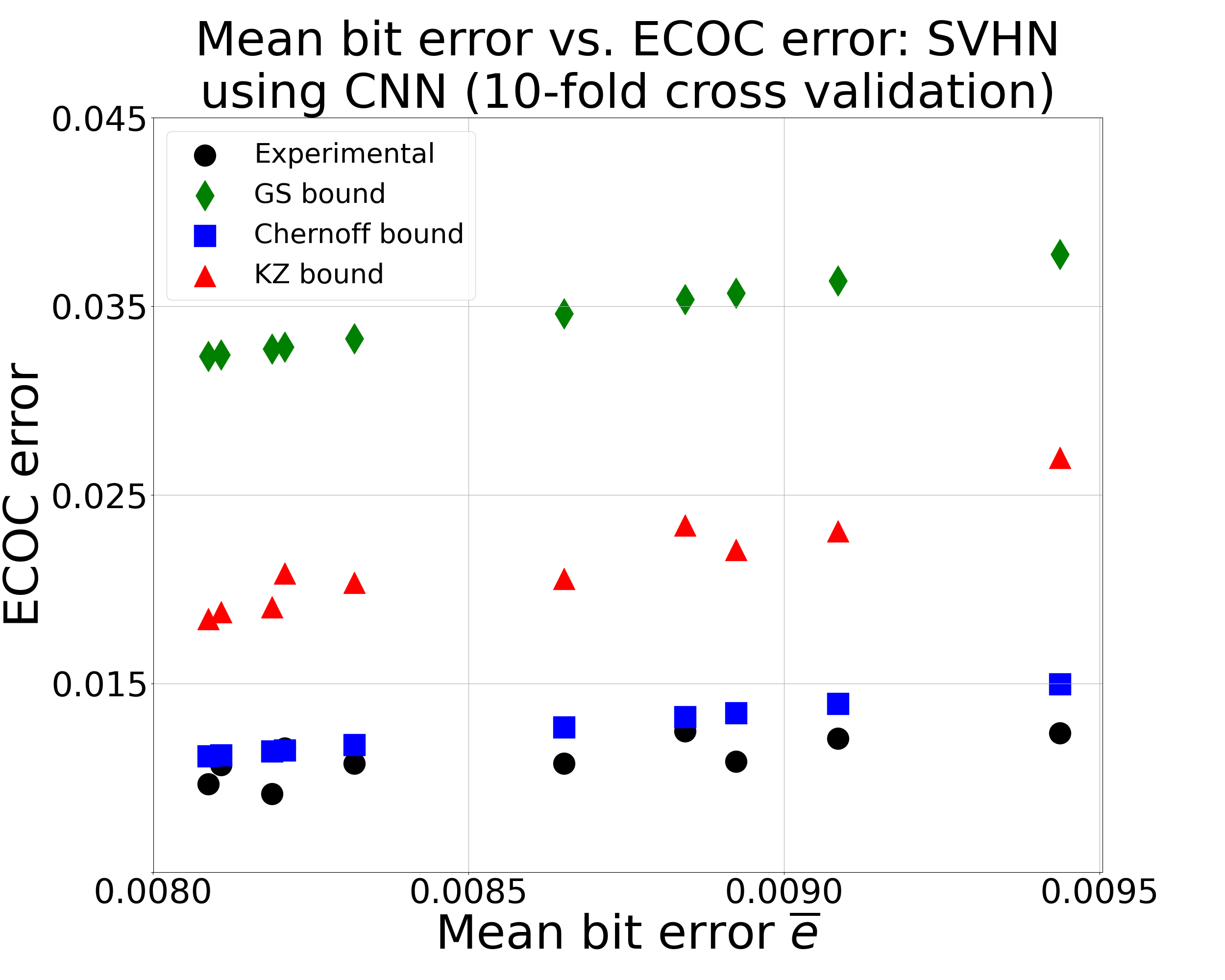}
\caption{SVHN: Mean bit error vs ECOC \\ error (CNN using 10-fold cross-validation)}
\label{fig:cifar-svm}
\end{figure}

\begin{table*}[htbp]
\begin{center}
    \begin{tabular}{|c|c|c|c|c|c|} \hline

 &  \multicolumn{5}{c|}{ECOC Error Rate} \\ \cline{2-6}  Dataset & Model & Experimental & GS &  Chernoff Bound & KZ Bound \\
\hline
Pendigits & DT & \textbf{0.034} $\pm$ 0.0034 &  0.134 $\pm$ 0.0070  & 0.148 $\pm$ 0.0130  & 0.192 $\pm$ 0.03450   \\ \cline{2-6}
 & SVM & \textbf{0.022} $\pm$ 0.0024 & 0.047 $\pm$ 0.0059 & 0.023 $\pm$ 0.0054 & 0.030 $\pm$ 0.0071  \\ \hline
Usps & DT &  \textbf{0.091} $\pm$ 0.0117  &  0.288 $\pm$ 0.0209  & 0.466 $\pm$ 0.0431  &  0.500 $\pm$ 0.0482   \\ \cline{2-6}
& SVM &  \textbf{0.028} $\pm$ 0.0050 &  0.063 $\pm$ 0.0085  & 0.040 $\pm$ 0.0100  &  0.049 $\pm$ 0.0149   \\ \hline
Vowel & DT  &  \textbf{0.144} $\pm$ 0.0397  &  0.449 $\pm$ 0.0604  & 0.749 $\pm$ 0.0833  &  0.746 $\pm$ 0.0626    \\ \cline{2-6}
& SVM &  \textbf{0.166} $\pm$ 0.0368 &  0.422 $\pm$ 0.0553  & 0.710 $\pm$ 0.0891  &  0.712 $\pm$ 0.0876   \\ \hline
Letters & DT &  0.061 $\pm$ 0.0057  &  0.274 $\pm$ 0.0114  & \textbf{0.047} $\pm$ 0.0082  &  0.055 $\pm$ 0.0108    \\
\cline{2-6}
& SVM &  0.106 $\pm$ 0.0046 &  0.302 $\pm$ 0.0086  & \textbf{0.070} $\pm$ 0.0081  &  0.093 $\pm$ 0.0191   \\ \hline
CIFAR-10 & CNN &  \textbf{0.023} $\pm$ 0.0015 &  0.065 $\pm$ 0.0042  & 0.041 $\pm$ 0.0049  &  0.074 $\pm$ 0.0098 \\ \hline
SVHN & CNN &  \textbf{0.011} $\pm$ 0.0010 & 0.034 $\pm$ 0.0018 & 0.013 $\pm$ 0.0013 & 0.021 $\pm$ 0.0025 \\ \hline
    \end{tabular}
 \captionsetup{justification=centering}
 \caption{\label{tab:error-rate} ECOC error rate (Mean and standard deviation of 10-fold cross-validation): \\ Experimental vs GS, Chernoff, and KZ bounds.}

\end{center}
\end{table*}

\section{Conclusions and Future Works}
In this paper, we presented two new classification error bounds for the ECOC ensemble learning: the first under the assumption that all base classifiers are independent and the second under the assumption that all base classifiers are mutually correlated up to first-order. These bounds have exponential decay complexity with respect to codeword length and theoretically validate the effectiveness of the ECOC approach. Moreover, we perform ECOC classification on six datasets and compare their error rates with our bounds to experimentally validate our work and show the effect of correlation on classification accuracy. Future work include investigating the Chernoff bound for ECOC in settings with limited independence \cite{schmidt1995} and comparing the performance of binary vs $N$-ary ECOC with respect to the error bounds presented in this paper.


\bibliographystyle{IEEEtran}
\bibliography{ictai2021}

\begin{thebibliography}{10}
\providecommand{\url}[1]{#1}
\csname url@samestyle\endcsname
\providecommand{\newblock}{\relax}
\providecommand{\bibinfo}[2]{#2}
\providecommand{\BIBentrySTDinterwordspacing}{\spaceskip=0pt\relax}
\providecommand{\BIBentryALTinterwordstretchfactor}{4}
\providecommand{\BIBentryALTinterwordspacing}{\spaceskip=\fontdimen2\font plus
\BIBentryALTinterwordstretchfactor\fontdimen3\font minus
  \fontdimen4\font\relax}
\providecommand{\BIBforeignlanguage}[2]{{%
\expandafter\ifx\csname l@#1\endcsname\relax
\typeout{** WARNING: IEEEtran.bst: No hyphenation pattern has been}%
\typeout{** loaded for the language `#1'. Using the pattern for}%
\typeout{** the default language instead.}%
\else
\language=\csname l@#1\endcsname
\fi
#2}}
\providecommand{\BIBdecl}{\relax}
\BIBdecl

\bibitem{db}
T.~Dietterich and G.~Bakiri, ``Solving multiclass learning problems via
  error-correcting output codes,'' \emph{J. Artificial Intelligence Research},
  vol.~2, pp. 263--286, January 1995.

\bibitem{escalera2008decoding}
S.~Escalera, O.~Pujol, and P.~Radeva, ``On the decoding process in ternary
  error-correcting output codes,'' \emph{IEEE transactions on pattern analysis
  and machine intelligence}, vol.~32, no.~1, pp. 120--134, 2008.

\bibitem{zhou2019}
J.~T. Zhou, I.~W. Tsang, S.~Ho, and K.~M\"{u}ller, ``\textit{N}-ary
  decomposition for multi-class classification,'' \emph{Mach. Learn.}, vol.
  108, pp. 809--830, February 2019.

\bibitem{liu2015joint}
M.~Liu, D.~Zhang, S.~Chen, and H.~Xue, ``Joint binary classifier learning for
  ecoc-based multi-class classification,'' \emph{IEEE Transactions on Pattern
  Analysis and Machine Intelligence}, vol.~38, no.~11, pp. 2335--2341, 2015.

\bibitem{zhong2013adaptive}
G.~Zhong and M.~Cheriet, ``Adaptive error-correcting output codes,'' in
  \emph{Twenty-Third International Joint Conference on Artificial
  Intelligence}, 2013.

\bibitem{passerini2004new}
A.~Passerini, M.~Pontil, and P.~Frasconi, ``New results on error correcting
  output codes of kernel machines,'' \emph{IEEE transactions on neural
  networks}, vol.~15, no.~1, pp. 45--54, 2004.

\bibitem{qin2017zero}
J.~Qin, L.~Liu, L.~Shao, F.~Shen, B.~Ni, J.~Chen, and Y.~Wang, ``Zero-shot
  action recognition with error-correcting output codes,'' in \emph{Proceedings
  of the IEEE Conference on Computer Vision and Pattern Recognition}, 2017, pp.
  2833--2842.

\bibitem{ho2020error}
S.-S. Ho, M.~Marchiano, S.~Zockoll, and H.~Nguyen, ``An error-correcting output
  code framework for lifelong learning without a teacher,'' in \emph{2020 IEEE
  32nd International Conference on Tools with Artificial Intelligence
  (ICTAI)}.\hskip 1em plus 0.5em minus 0.4em\relax IEEE, 2020, pp. 249--254.

\bibitem{song2021error}
Y.~Song, Q.~Kang, W.~P. Tay, Y.~Song, Q.~Kang, and W.~Tay, ``Error-correcting
  output codes with ensemble diversity for robust learning in neural
  networks,'' in \emph{Proceedings of the AAAI Conference on Artificial
  Intelligence}, vol.~35, no.~11, 2021, pp. 9722--9729.

\bibitem{gs}
V.~Guruswami and A.~Sahai, ``Multiclass learning, boosting, and
  error-correcting codes,'' in \emph{COLT 1999}, 1999, pp. 145--155.

\bibitem{kz}
S.~Kaniovski and A.~Zaigraev, ``Optimal jury design for homogeneous juries with
  correlated votes,'' \emph{Theory and Decision}, vol.~71, pp. 439--459, 2011.

\bibitem{allwein}
E.~L. Allwein, R.~E. Schapire, and Y.~Singer, ``Reducing multiclass to binary:
  A unifying approach for margin classifiers,'' \emph{J. Mach. Learn.}, vol.~1,
  pp. 113--141, 2001.

\bibitem{kjo}
A.~Klautau, N.~Jevti\'{c}, and A.~Orlitsky, ``On nearest-neighbor
  error-correcting output codes with application to all-pairs multiclass
  support vector machines,'' \emph{J. Mach. Learn. Research}, vol.~4, pp.
  1--15, 2003.

\bibitem{langford2005}
J.~Langford and A.~Beygelzimer, ``Sensitive error correcting output codes,'' in
  \emph{COLT 2005}, 2005, pp. 158--172.

\bibitem{beygelzimer2009}
A.~Beygelzimer, J.~Langford, and P.~Ravikumar, ``Error-correcting
  tournaments,'' in \emph{International Conference on Algorithmic Learning
  Theory (ALT), 2009}, 2009, pp. 247--262.

\bibitem{blum1998}
A.~Blum and T.~Mitchell, ``Combining labeled and unlabeled data with
  co-training,'' in \emph{COLT 1998}, 1998, pp. 92--100.

\bibitem{dasgupta2001}
S.~Dasgupta, M.~L. Littman, and D.~A. McAllester, ``Sensitive error correcting
  output codes,'' in \emph{NIPS 2005}, 2001, pp. 375--382.

\bibitem{balcan2004}
M.-F. Balcan, A.~Blum, and K.~Yang, ``Co-training and expansion: towards
  bridging theory and practice,'' in \emph{NIPS 2004}, 2004, pp. 89--96.

\bibitem{nguyen2021appendix}
H.~D. Nguyen, M.~S. Khan, N.~Kaegi, S.-S. Ho, J.~Moore, L.~Borys, and
  L.~Lavalva, ``Ensemble learning using error correcting output codes: New
  classification error bounds: Appendix,''
  \url{https://drive.google.com/file/d/1SuqVu2q9GFazV8FPfBEFT99ItcQM-GF3/view},
  2021.

\bibitem{feller}
W.~Feller, \emph{An Introduction to Probability Theory and its
  Applications}.\hskip 1em plus 0.5em minus 0.4em\relax John Wiley and Sons,
  1968, vol.~1.

\bibitem{chernoff}
H.~Chernoff, ``A measure of asymptotic efficiency for tests of a hypothesis
  based on the sum of observations,'' \emph{The Annals of Mathematical
  Statistics}, vol.~23, pp. 493--507, 1952.

\bibitem{osg07}
R.~Pordes, D.~Petravick, B.~Kramer, D.~Olson, M.~Livny, A.~Roy, P.~Avery,
  K.~Blackburn, T.~Wenaus, F.~W{\"u}rthwein, I.~Foster, R.~Gardner, M.~Wilde,
  A.~Blatecky, J.~McGee, and R.~Quick, ``The open science grid,'' in \emph{J.
  Phys. Conf. Ser.}, ser. 78, vol.~78, 2007, p. 012057.

\bibitem{osg09}
I.~Sfiligoi, D.~C. Bradley, B.~Holzman, P.~Mhashilkar, S.~Padhi, and
  F.~Wurthwein, ``The pilot way to grid resources using glideinwms,'' in
  \emph{2009 WRI World Congress on Computer Science and Information
  Engineering}, ser. 2, vol.~2, 2009, pp. 428--432.

\bibitem{schmidt1995}
J.~P. Schmidt, A.~Siegel, and A.~Srinivsan, ``Chernoff-hoeffding bounds for
  applications with limited independence,'' \emph{SIAM Journal on Discrete
  Mathematics}, vol.~8, no.~2, pp. 223--250, 1995.

\end{thebibliography}


\pagebreak
\appendix


In this appendix we present proofs of lemmas and theorems that were not included in our paper and additional expperimental results.

\subsection{Proofs of lemmas and theorems}

\begin{proof}[Proof of Lemma \ref{le:inequality}] It suffices to prove (\ref{eq:inequality}) only for $m=n$ because of symmetry.  Observe that
 \[
 \frac{|S(n-1,k)|}{|S(n-1,k-1)|} = \frac{\binom{n-1}{k}}{\binom{n-1}{k-1}} = \frac{n-k}{k},
 \]
or equivalently,
\begin{equation} \label{eq:card}
k|S(n-1,k)|=(n-k)|S(n-1,k-1)|.
\end{equation}
 
Define $S_m(n,k)$ to be the multi-set consisting of the union of $m$ copies of $S(n,k)$.  Then $|S_k(n-1,k)|=|S_{n-k}(n-1,k-1)|$ because of (\ref{eq:card}).  We now differentiate and rearrange terms as follows:
 
\begin{align*}
& \frac{dp_E(n,k)}{de_n} \\
&=  \sum_{\substack{A\in S(n-1,k-1)}} \left(\prod_{i\in A} e_i\right)\left(\prod_{j\in \bar{A}} (1-e_j)\right) \\
& \ \ - \sum_{\substack{A\in S(n-1,k)}} \left(\prod_{i\in A} e_i\right)\left(\prod_{j\in \bar{A}} (1-e_j)\right) \\
& = \sum_{\substack{A\in S_{n-k}(n-1,k-1)}} \frac{1}{n-k} \left(\prod_{i\in A} e_i\right)\left(\prod_{j\in \bar{A}} (1-e_j)\right)  \\
& \ \ - \sum_{\substack{A\in S_k(n-1,k)}} \frac{1}{k} \left(\prod_{i\in A} e_i\right)\left(\prod_{j\in \bar{A}} (1-e_j)\right) \\
& = \sum_{\substack{A\in S(n-1,k-1)}}\left[ \sum_{l\in \bar{A}} \frac{1}{n-k} \left(\prod_{i\in A} e_i\right) 
\left(\prod_{j\in \bar{A}} (1-e_j)\right)\right]  \\
& \ \  - \sum_{\substack{A\in S(n-1,k-1)}} \left[ \sum_{l\in \bar{A}}  \frac{1}{k} \left(\prod_{i\in A\cup \{l\}} e_i\right)\left(\prod_{j\in \bar{A}-\{l\}} (1-e_j)\right) \right] \\
& = \sum_{\substack{A\in S(n-1,k-1)}}\left[ \sum_{l\in \bar{A}} \left(\prod_{i\in A} e_i\right) 
\left(\prod_{j\in \bar{A}-\{l\}} (1-e_j)\right) \right. \\
& \ \ \ \ \left. \left(\frac{1-e_l}{n-k}-\frac{e_l}{k}\right) \right] \\
& >0
\end{align*}
since
\[
\frac{1-e_l}{n-k}-\frac{e_l}{k}>0, \ \ i=1,\ldots, n,
\]
which follows from the assumption that every $e_l < k/n$.  
\end{proof}

\begin{proof} [of Lemma \ref{le:monotone}] Observe that $p_E(n,k)$ is linear in each $e_m$.  Assume $k\geq 1$.  We claim that
 \begin{equation}\label{eq:dp}
 \frac{dp_E(n,k)}{de_m}=p_{E_m}(n-1,k-1) - p_{E_m}(n-1,k) 
 \end{equation}
 for all $m \in [n]$.  By symmetry, it suffices to prove (\ref{eq:dp}) for $m=n$.  We have
\begin{align*}
\frac{dp_E(n,k)}{de_n} & = \frac{d}{de_n}\left[ \sum_{A\in S(n,k)} \left(\prod_{i\in A} e_i\right)\left(\prod_{j\in \bar{A}} (1-e_j)\right) \right] \\
& = \frac{d}{de_n}\left[ \sum_{\substack{A\in S(n,k)\\  n \in A}} \left(\prod_{i\in A} e_i\right)\left(\prod_{j\in \bar{A}} (1-e_j)\right) \right. \\
& \ \ \ \ \left. + \sum_{\substack{A\in S(n,k)\\ n \notin A}} \left(\prod_{i\in A} e_i\right)\left(\prod_{j\in \bar{A}} (1-e_j)\right) \right] \\
& =  \sum_{\substack{A\in S(n-1,k-1)}} \left(\prod_{i\in A} e_i\right)\left(\prod_{j\in \bar{A}} (1-e_j)\right) \\
& \ \ \ \  - \sum_{\substack{A\in S(n-1,k)}} \left(\prod_{i\in A} e_i\right)\left(\prod_{j\in \bar{A}} (1-e_j)\right) \\
& =p_{E_n}(n-1,k-1) - p_{E_n}(n-1,k) 
\end{align*}

By the previous lemma, we have
 \[p_{E_m}(n-1,k-1) - p_{E_m}(n-1,k) >0
 \]
when $0\leq e_i < k/n$ for every $i\in A_m$, which proves that $\frac{dp_E(n,k)}{de_m} >0$.  Thus, $p_E(n,k)$ is strictly increasing with respect to $e_n$, and therefore strictly increasing with respect to each $e_m$ due to symmetry.
 \end{proof}

\section{Proofs of Lemma \ref{le:pcor} and \ref{le:pincreasing}}

\begin{proof}[of Lemma \ref{le:pcor}]
We first assume the case where $k\leq n-2$ and consider the following partition of $S(n,k)$:
\[
S(n,k)=S(n-2,k)\cup S'\cup S'',
\]
where
\begin{align}
S' & = \{A'=A\cup\{n-1\}:A\in S(n-2,k-1)\} \notag \\
& \ \ \ \ \cup \{A'=A\cup\{n\}: A\in S(n-2,k-1)\} \\
S''& =\{A''=A\cup\{n-1,n\}:A\in S(n-2,k-2)\}
\end{align}
Then
\begin{align*}
p_{E_n}(n,k,f) & = \sum_{A\in S(n,k)} P_n(A)   \\
& =\sum_{A\in S(n-2,k)} P_n(A) + \sum_{A'\in S'} P_n(A)  \\
& \ \ \ \ + \sum_{A''\in S''} P_n(A) \\
& = (1-e_{n-1}-e_n + f)\sum_{A\in S(n-2,k)} P_{n-2}(A) \\
& \ \ \ \ + (e_{n-1}-f) \sum_{A\in S(n-2,k-1)} P_{n-2}(A) \\
& \ \ \ \ + (e_n-f) \sum_{A\in S(n-2,k-1)} P_{n-2}(A) \\
& \ \ \ \ + f\sum_{A\in S(n-2,k-2)} P_{n-2}(A) \\
& = (1-2\bar{e}_n + f)p_{E_{n-2}}(n-2,k) \\
& \ \ \ \ + 2(\bar{e}_n-f) p_{E_{n-2}}(n-2,k-1) \\
& \ \ \ \ + f p_{E_{n-2}}(n-2,k-2)
\end{align*}
This proves (\ref{eq:pcor}) for this case.  The other two cases, $k=n-1$ and $k=n$, can be proven by a similar argument by considering appropriate partitions of $S(n,n-1)$ and $S(n,n)$, respectively.
\end{proof}

\begin{proof}[Proof of Lemma \ref{le:pincreasing}]
We first assume the case where $k\leq n-2$.  It suffices to prove that the derivative of $p(n,k,e,f)$ with respect to $f$ is non-negative.  Since Lemma \ref{le:pcor} shows that $p(n,k,e,f)$ is linear in $f$, we have
\begin{align}
\frac{dp(n,k,\bar{e},f)}{df} & = p(n-2,k-2,\bar{e})-2p(n-2,k-1,\bar{e}) \notag \\
& \ \ \ \ + p(n-2,k,\bar{e}) \label{eq:dp1} \\
& = \binom{n-2}{k-2}\bar{e}^{k-2}(1-\bar{e})^{n-k} \\
& \ \ \ \ -2\binom{n-2}{k-1}\bar{e}^{k-1}(1-\bar{e})^{n-k-1} \\ 
& \ \ \ \ +\binom{n-2}{k}\bar{e}^k(1-\bar{e})^{n-k-2} \\
& = \bar{e}^{k-2}(1-\bar{e})^{n-k-2}\binom{n-2}{k-2} \left[ \frac{R(\bar{e})}{k(k-1)}\right],
\end{align}
where
\[
R(\bar{e})=(1-\bar{e})^2k(k-1)-2\bar{e}(1-\bar{e})k(n-k)+\bar{e}^2(n-k)(n-k-1).
\]
Solving $R(\bar{e})=0$ for $\bar{e}$ yields
\[
\bar{e}=\frac{k}{n} \pm \frac{1}{n}\sqrt{\frac{k(n-k)}{n-1}}=\frac{k}{n}\pm \alpha(n,k).
\]
It is straightforward to verify that $R(\bar{e})\geq 0$ for  $0\leq \bar{e} \leq k/n- \alpha(n,k)$ and therefore, $\frac{dp(n,k,\bar{e},f)}{df} \geq 0$ over the same domain.  This proves that $p(n,k,\bar{e},f)$ is increasing with respect to $f$. 

As for the case where $k=n-1$, we have
\begin{align}
\frac{dp(n,n-1,\bar{e},f)}{df} & = p(n-2,n-3,\bar{e})-2p(n-2,n-2,\bar{e})  \label{eq:dp2}\\
& = \binom{n-2}{n-3}\bar{e}^{n-3}(1-\bar{e}) -2\binom{n-2}{n-2}\bar{e}^{n-2} \\
& = \bar{e}^{n-3}[(n-2)(1-\bar{e})-2\bar{e}] \\
& = \bar{e}^{n-3}[n(1-\bar{e})-2]
\end{align}
It follows that $\frac{dp(n,n-1,\bar{e},f)}{df} \geq 0$ when $0\leq \bar{e} \leq 1-2/n$ and proves that $p(n,n-1,\bar{e},f)$ is increasing with respect to $f$. 

Lastly, in the case where $k=n$, we have
As for the case where $k=n-1$, we have
\begin{equation} \label{eq:dp3}
\frac{dp(n,n-1,\bar{e},f)}{df} = p(n-2,n-2,\bar{e})
\end{equation}
which is clearly non-negative for all $\bar{e}\in [0,1]$ and proves that $p(n,n,\bar{e},f)$ is increasing with respect to $f$. 
\end{proof}

\section{Proofs of Theorems \ref{th:epsilon-correlation} and \ref{th:correlation-monotone}}

\begin{proof}[of Theorem \ref{th:epsilon-correlation}]
We have
\begin{align*}
\varepsilon(n,m,\bar{e},f) & = \sum_{k=m}^n p(n,k,\bar{e},f) \\
&  = \sum_{k=m}^{n-2} p(n,k,\bar{e},f) +  p(n,n-1,\bar{e},f) \\
& \ \ \ \ +  p(n,n,\bar{e},f) \\
& =\sum_{k=m}^{n-2} [fp(n-2,k-2,\bar{e}) \\
& \ \ \ \ +2(\bar{e}-f)p(n-2,k-1,\bar{e}) \\
& \ \ \ \ + (1-2\bar{e}+f)p(n-2,k,\bar{e}) ]  \\
& \ \ \ \ + [fp(n-2,n-3,\bar{e}) \\
& \ \ \ \ +2(\bar{e}-f)p(n-2,n-2,\bar{e})] \\
& \ \ \ \ + fp(n-2,n-2,\bar{e})  \\
& = f \cdot \sum_{k=m-2}^{n-2}p(n-2,k,\bar{e}) \\ 
& \ \ \ \ +  2(\bar{e}-f) \cdot \sum_{k=m-1}^{n-2}p(n-2,k,\bar{e}) \\
& \ \ \ \ +  (1-2\bar{e}+f) \cdot \sum_{k=m}^{n-2}p(n-2,k,\bar{e}) \\
& = f\varepsilon(n-2,m-2,\bar{e}) \\
& \ \ \ \ +2(\bar{e}-f)\varepsilon(n-2,m-1,\bar{e}) \\
& \ \ \ \ +(1-2\bar{e}+f)\varepsilon(n-2,m,\bar{e})
\end{align*}

\end{proof}

\begin{proof}[Proof of Theorem \ref{th:correlation-monotone}]
We compute the derivative of $\varepsilon(n,m,\bar{e},f)$ with respect to $f$ using identities (\ref{eq:dp1}), (\ref{eq:dp2}), and (\ref{eq:dp3}), and the fact that the following sum telescopes:

\begin{align}
& \frac{d\varepsilon(n,m,\bar{e},f)}{df} \notag \\
& =\sum_{k=m}^n \frac{dp(n,k,\bar{e},f)}{df} \\
&  =\sum_{k=m}^{n-2} \frac{dp(n,k,\bar{e},f)}{df} +  \frac{dp(n,n-1,\bar{e},f)}{df} \notag \\
& \ \ \ \ +  \frac{dp(n,n,\bar{e},f)}{df} \\
 & =\sum_{k=m}^{n-2} [p(n-2,k-2,\bar{e})-2p(n-2,k-1,\bar{e}) \notag \\
 & \ \ \ \ + p(n-2,k,\bar{e}) ]
 +  [p(n-2,n-3,\bar{e}) \notag \\
 & \ \ \ \ -2p(n-2,n-2,\bar{e})] + p(n-2,n-2,\bar{e}) \\
 & = [p(n-2,m-2,\bar{e})-2p(n-2,m-1,\bar{e}) \\
 & \ \ \ \ + p(n-2,m,\bar{e}) ]
+ [p(n-2,m-1,\bar{e}) \\
& \ \ \ \ -2p(n-2,m,\bar{e})+ p(n-2,m+1,\bar{e}) ] \\
  & \ \ \ \ + [p(n-2,m,\bar{e})-2p(n-2,m+1,\bar{e}) \\
  & \ \ \ \ + p(n-2,m+2,\bar{e}) ] \\
  & \ \ \ \ ... \\
   & \ \ \ \ + [p(n-2,n-4,\bar{e})-2p(n-2,n-3,\bar{e}) \\
   & \ \ \ \ + p(n-2,n-2,\bar{e}) ] \\
 & \ \ \ \ +  [p(n-2,n-3,\bar{e})-2p(n-2,n-2,\bar{e})] \\
 & \ \ \ \ + p(n-2,n-2,\bar{e}) \\
 & = p(n-2,m-2,\bar{e})-p(n-2,m-1,\bar{e})
\end{align}
It follows that $\frac{d\varepsilon(n,m,\bar{e},f)}{df} \geq 0$ when $ p(n-2,m-2,\bar{e})-p(n-2,m-1,\bar{e}) \geq 0$, or equivalently,
\begin{align}
 p(n-2,m-2,\bar{e}) & \geq p(n-2,m-1,\bar{e}) \\
\Rightarrow \binom{n-2}{m-2}\bar{e}^{m-2}(1-\bar{e})^{n-m} & \geq \binom{n-2}{m-1}\bar{e}^{m-1}(1-\bar{e})^{n-m-1} \\
\therefore \bar{e} & \leq \frac{m-1}{n-1} 
\end{align}
Thus, $p(n,m,\bar{e},f)$ is increasing with respect to $f$ for $0\leq \bar{e} \leq \frac{m-1}{n-1}$.  By the same argument, $p(n,m,\bar{e},f)$ must be decreasing with respect to $f$ for $\frac{m-1}{n-1} \leq \bar{e} \leq 1$.
\end{proof}

\section{Proof of Theorem \ref{th:kz}}

\begin{proof}[Proof of Theorem \ref{th:kz}]
The probability $\pi_A$ can be expressed in terms of $P(A)=\bar{e}^k(1-\bar{e})^{n-k}$, the corresponding probability assuming zero correlation, by
\begin{align}
\pi_A & = P(A) \left(1+\sum_{i<j} c_{ij} z_iz_j \right) \\
& = P(A) \left(1+c\sum_{i<j} \frac{L_i-e_i}{\sqrt{e_i(1 - e_i)}}\frac{L_j-e_j}{\sqrt{e_j(1 - e_j)}} \right) \\
& = P(A) \left(1+\frac{c}{\bar{e}(1-\bar{e})}\sum_{i<j} (L_i-\bar{e})(L_j-\bar{e}) \right) \\
& =P(A) \left(1+\frac{c}{2\bar{e}(1-\bar{e})}\sum_{i \neq j} (L_i-\bar{e})(L_j-\bar{e}) \right) \\
& = P(A) \left(1+\frac{c}{2\bar{e}(1-\bar{e})}\left(\sum_{i,j} (L_i-\bar{e})(L_j-\bar{e}) \right. \right. \notag \\
& \ \ \ \ \left. \left. -\sum_{i=j} (L_i-\bar{e})(L_j-\bar{e}) \right) \right) \\
& = P(A) \left(1+\frac{c}{2\bar{e}(1-\bar{e})}\left(\sum_{i,j} (L_iL_j-\bar{e}(L_i+L_j)+\bar{e}^2) \right. \right. \notag \\
& \ \ \ \ \left. \left. -\sum_{i=j} (L_i-\bar{e})(L_j-\bar{e}) \right) \right) \\
& = P(A) \left(1+\frac{c}{2\bar{e}(1-\bar{e})}\left((\sum_{i}^n L_i)^2 -\bar{e} \sum_{i,j}(L_i+L_j) \right. \right. \notag \\
& \ \ \ \ \left. \left. +\sum_{i,j}\bar{e}^2 -\sum_{i=j} (L_i-\bar{e})(L_j-\bar{e}) \right) \right) \\
& = P(A) \left(1+\frac{c}{2\bar{e}(1-\bar{e})}\left(k^2 -2\bar{e}nk+\bar{e}^2n^2  \right. \right. \notag \\
& \ \ \ \ \left. \left. -\sum_{i=j} (L_iL_j-\bar{e}(L_i+L_j)+\bar{e}^2) \right) \right) \\
& = P(A) \left(1+\frac{c}{2\bar{e}(1-\bar{e})}\left(k^2 -2\bar{e}nk+\bar{e}^2n^2 \right. \right. \notag \\
& \ \ \ \ \left. \left. -\sum_{i=j} L_iL_j + \sum_{i=j} \bar{e}(L_i+L_j) -\sum_{i=j} \bar{e}^2) \right) \right) \\
& = P(A) \left(1+\frac{c}{2\bar{e}(1-\bar{e})}\left(k^2 -2\bar{e}nk+\bar{e}^2n^2 -k  \right. \right. \notag \\
& \ \ \ \ \left. \left. + 2\bar{e}k -\bar{e}^2n) \right) \right) \\
& = P(A) \left(1+\frac{c}{2\bar{e}(1-\bar{e})}\left(k^2 - k + \bar{e}(n-1)(n\bar{e}-2k) \right) \right)
\end{align}
It follows that
\begin{align}
 \varepsilon(n,m,\bar{e},c)
& = \sum_{k=m}^n \sum_{A\in S(n,k)}\pi_A \\
& = \varepsilon(n,m,\bar{e}) \notag  \\
& \ \ \ \ + \frac{c}{2\bar{e}(1-\bar{e})}\sum_{k=m}^n \binom{n}{k}\bar{e}^k(1-\bar{e})^{n-k} \notag \\
& \ \ \ \ \ \ \ \ \left(k^2 - k + \bar{e}(n-1)(n\bar{e}-2k)\right)
\end{align}
The rest of the proof follows exactly as that given in \cite[Theorem 1]{kz}.
\end{proof}

\subsection{Additional Experimental Results}

In this section we provide  experimental results that were not included in our paper.  Figures 8-11 give plots of ECOC error rates for Usps and Vowels datasets.  Tables 3-8 give full results for each dataset, namely mean bit error, mean correlation, and ECOC error for all ten folds.

\begin{figure}

\includegraphics[width=240pt, height=165pt]{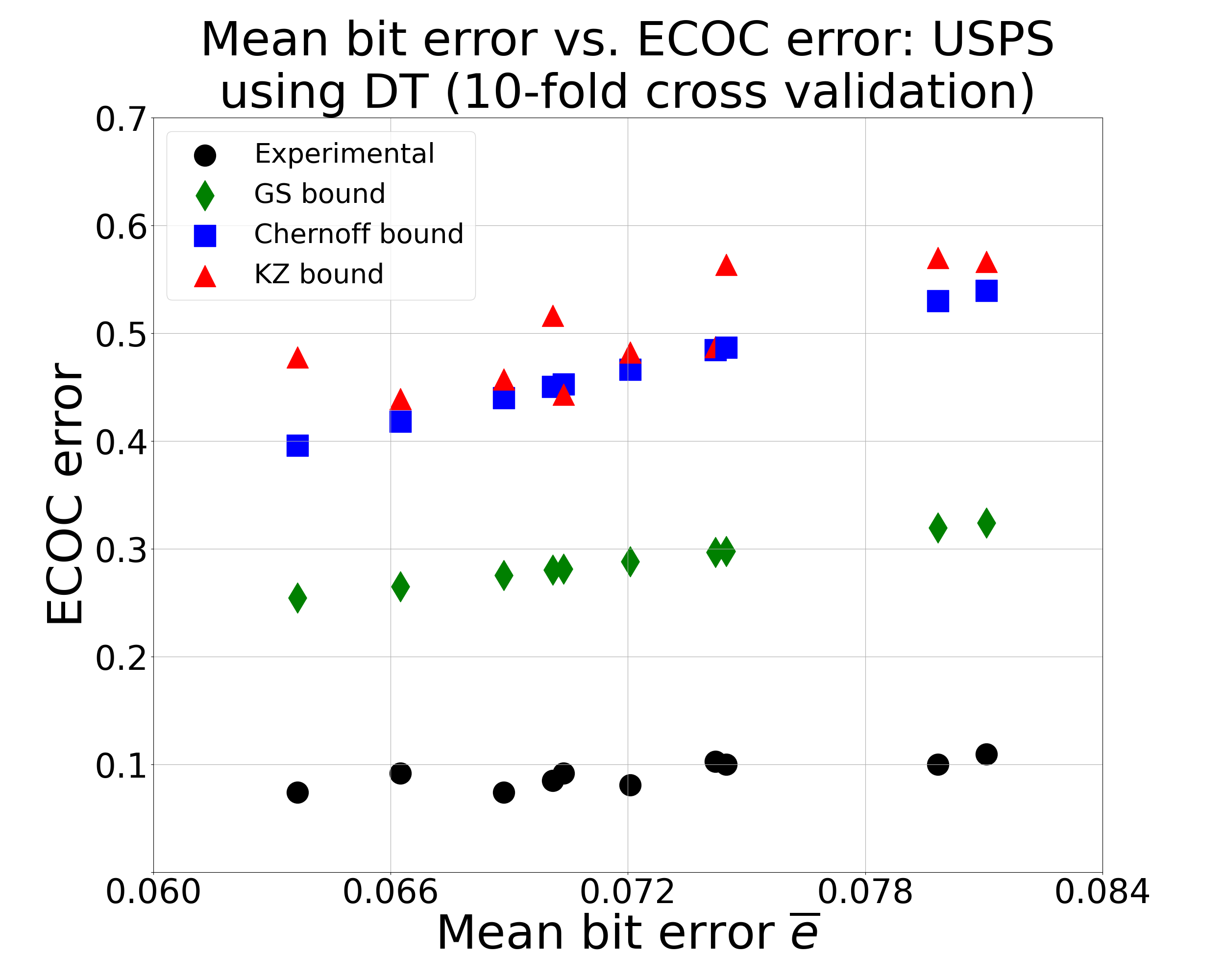}
\caption{USPS: Mean bit error vs ECOC \\ error (DT using 10-fold cross-validation)}
\label{fig:usps-dt}
\end{figure}

\begin{figure}
\includegraphics[width=240pt, height=165pt]{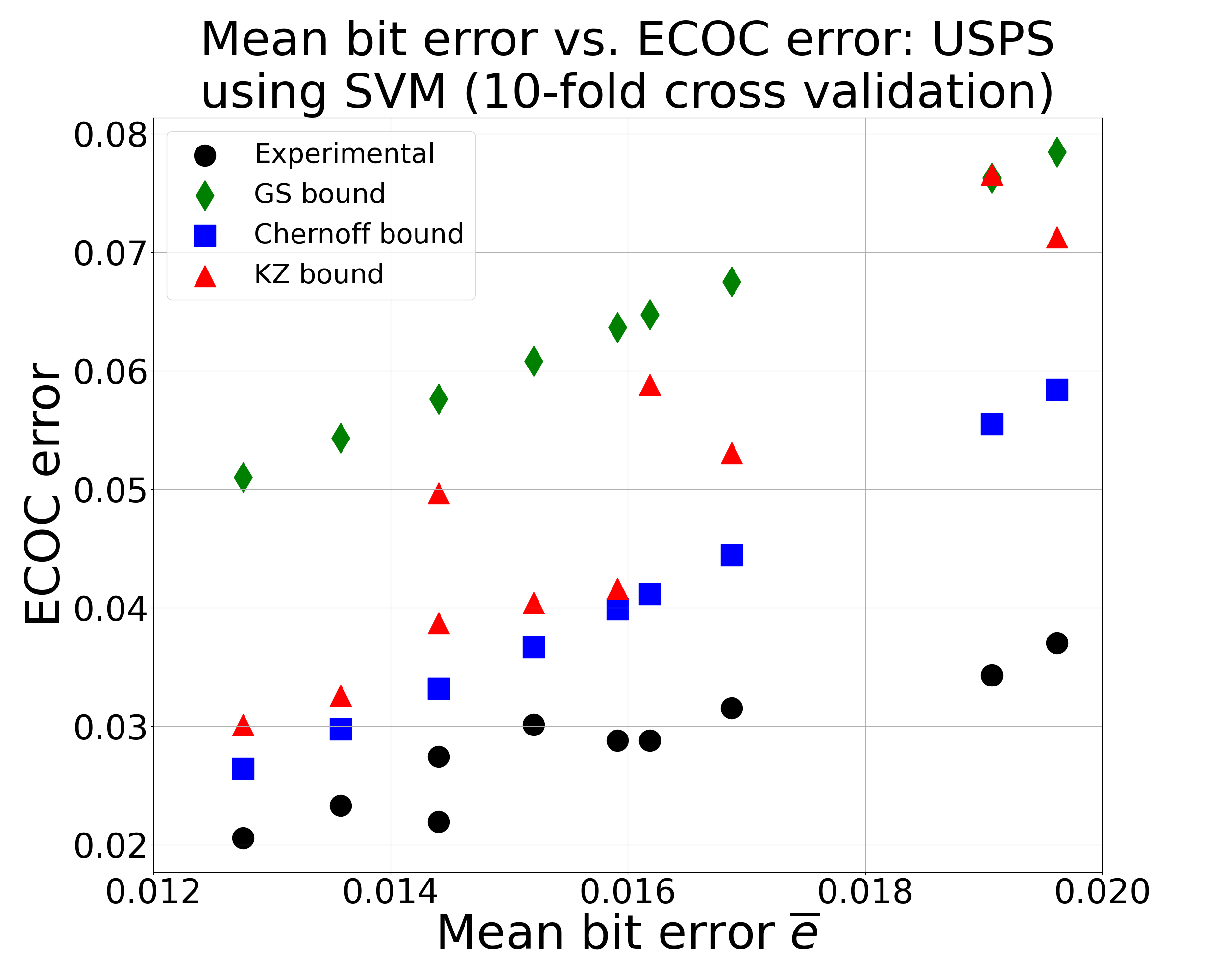}
\caption{USPS: Mean bit error vs ECOC \\ error (SVM using 10-fold cross-validation)}
\label{fig:usps-svm}
\end{figure}

\begin{figure}
\includegraphics[width=240pt, height=165pt]{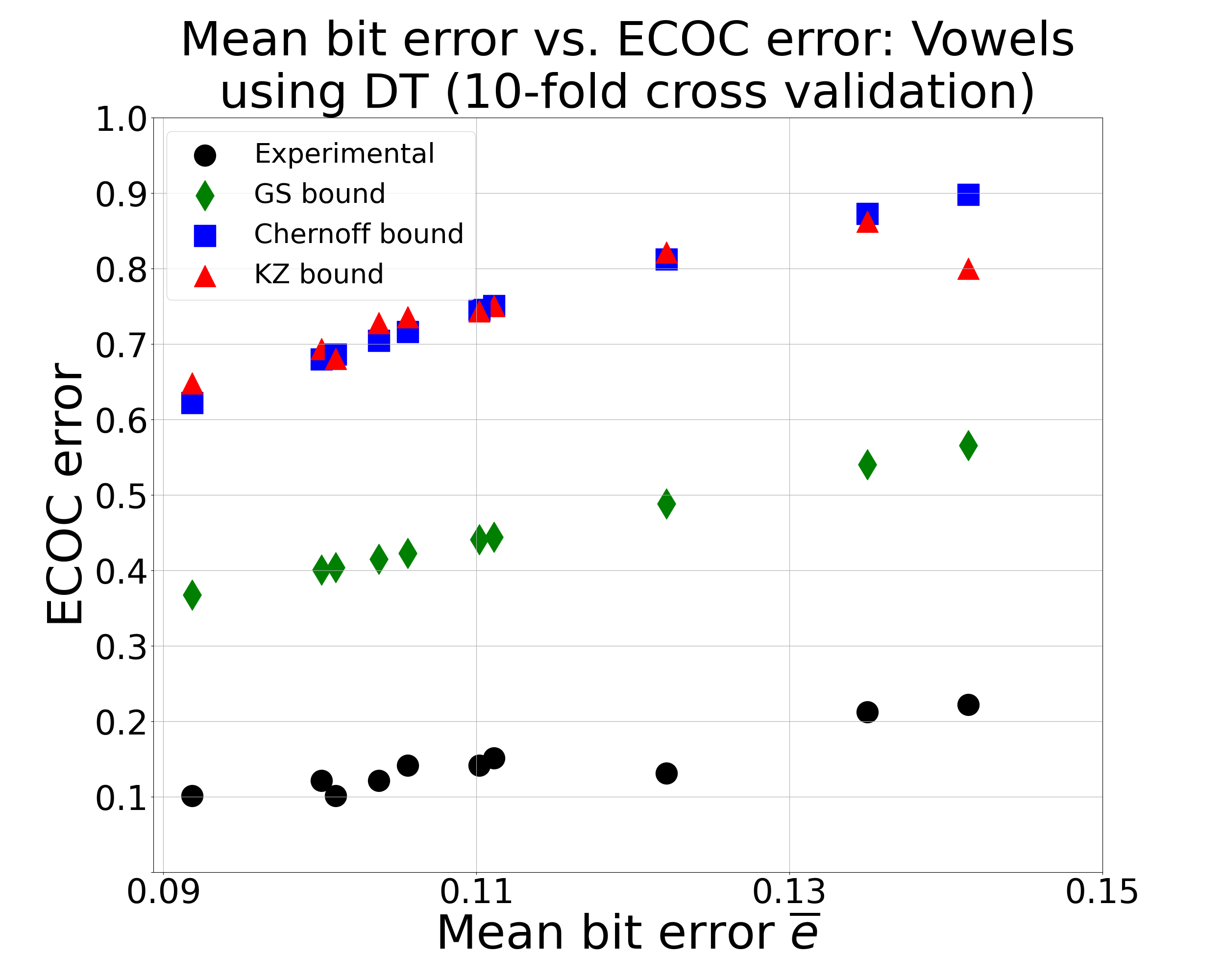}
\caption{Vowels: Mean bit error vs ECOC \\ error (DT using 10-fold cross-validation)}
\label{fig:vowels-dt}
\end{figure}

\begin{figure}
\includegraphics[width=210pt, height=150pt]{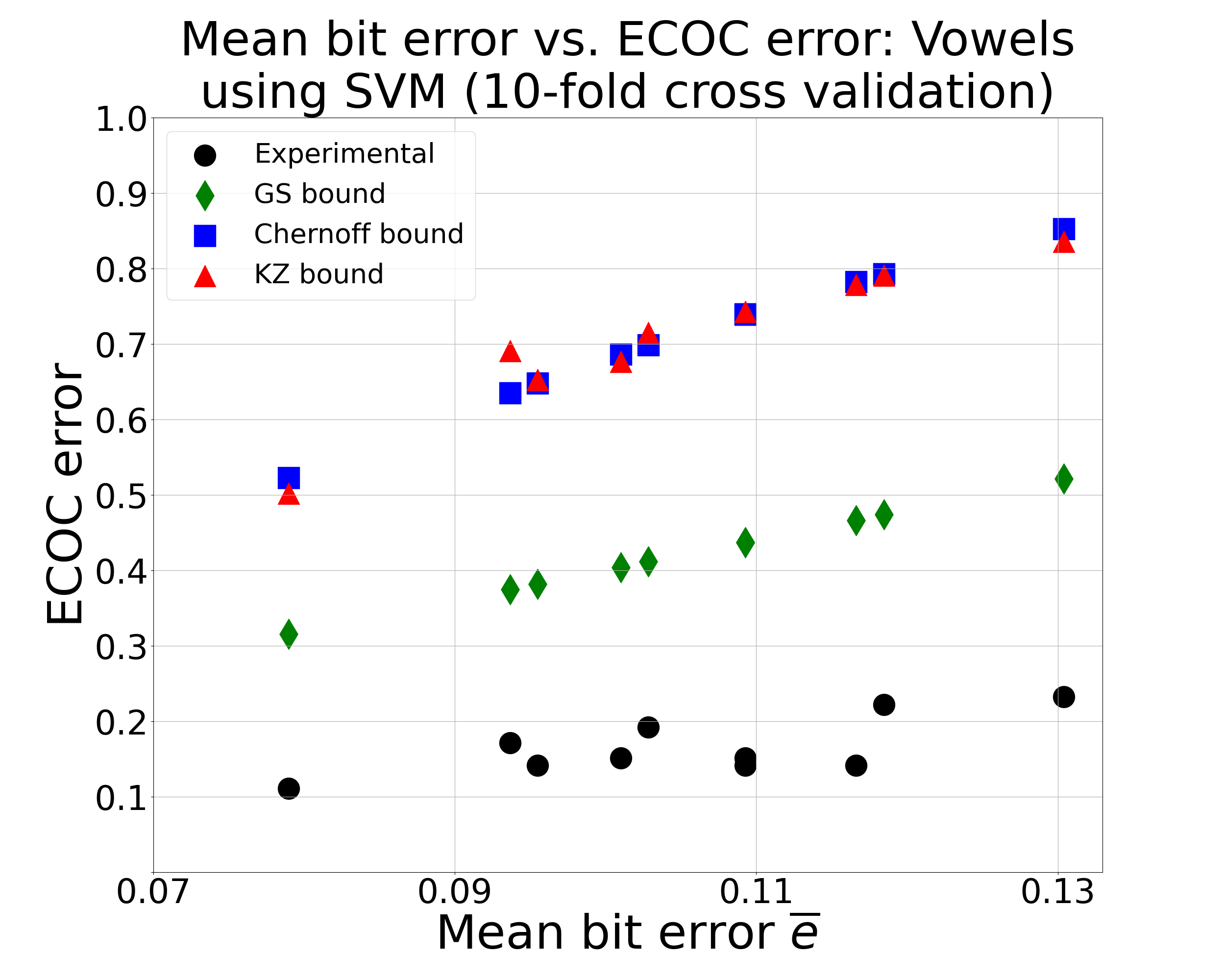}
\caption{Vowels: Mean bit error vs ECOC \\ error (SVM using 10-fold cross-validation)}
\label{fig:vowels-svm}
\end{figure}

\begin{table*}[htbp]
\begin{center}
    \begin{tabular}{|c|c|c|c|c|} \hline

  Fold & Model & Mean Bit Error & Mean Correlation & ECOC Error \\
\hline
1 & DT & 0.0323 $\pm$ 0.00915 & 0.0154 $\pm$ 0.34584 & 0.0328 \\
\cline{2-5}
 & SVM & 0.0136 $\pm$ 0.00380 & -0.0020 $\pm$ 0.34105 & 0.0237 \\
\hline
2 & DT & 0.0326 $\pm$ 0.00936 & 0.1356 $\pm$ 0.35876 & 0.0309 \\
\cline{2-5}
 & SVM & 0.0111 $\pm$ 0.00438 & 0.0956 $\pm$ 0.41435 & 0.0191 \\
\hline
3 & DT & 0.0315 $\pm$ 0.00747 & -0.0252 $\pm$ 0.3018 & 0.0328 \\
\cline{2-5}
 & SVM & 0.0094 $\pm$ 0.00288 & -0.0170 $\pm$ 0.28860 & 0.0218 \\
\hline
4 & DT & 0.0342 $\pm$ 0.00878 & 0.1666 $\pm$ 0.39814 & 0.0400 \\
\cline{2-5}
 & SVM & 0.0126 $\pm$ 0.00430 & 0.1234 $\pm$ 0.40851 & 0.0227 \\
\hline
5 & DT & 0.0313 $\pm$ 0.00809 & 0.0724 $\pm$ 0.33125 & 0.0336 \\
\cline{2-5}
 & SVM & 0.0105 $\pm$ 0.00273 & 0.0805 $\pm$ 0.36234 & 0.0191 \\
\hline
6 & DT & 0.0341 $\pm$ 0.00882 & 0.1196 $\pm$ 0.34530 & 0.0273 \\
\cline{2-5}
 & SVM & 0.0101 $\pm$ 0.00367 & 0.1119 $\pm$ 0.33386 & 0.0191 \\
\hline
7 & DT & 0.0332 $\pm$ 0.00817 & 0.1194 $\pm$ 0.34352 & 0.0328 \\
\cline{2-5}
 & SVM & 0.0120 $\pm$ 0.00357 & 0.0919 $\pm$ 0.35522 & 0.0255 \\
\hline
8 & DT & 0.0371 $\pm$ 0.01119 & 0.0637 $\pm$ 0.34278 & 0.0382 \\
\cline{2-5}
 & SVM & 0.0143 $\pm$ 0.00513 & 0.0150 $\pm$ 0.33852 & 0.0255 \\
\hline
9 & DT & 0.0326 $\pm$ 0.00631 & 0.1546 $\pm$ 0.41389 & 0.0328 \\
\cline{2-5}
 & SVM & 0.0120 $\pm$ 0.00415 & 0.1546 $\pm$ 0.41389 & 0.0209 \\
\hline
10 & DT & 0.0355 $\pm$ 0.00780 & 0.1576 $\pm$ 0.35864 & 0.0345 \\
\cline{2-5}
 & SVM & 0.0125 $\pm$ 0.00493 & 0.1111 $\pm$ 0.39133 & 0.0245 \\
\hline
    \end{tabular}
\caption{\label{tab:pendigigts-correlation} Pendigits: Mean bit error, mean correlation, ECOC error per fold}
\end{center}
\end{table*}

\begin{table*}[htbp]
\begin{center}
    \begin{tabular}{|c|c|c|c|c|} \hline

  Fold & Model & Mean Bit Error & Mean Correlation & ECOC Error \\
\hline
1 & DT & 0.0742 $\pm$ 0.01642 & 0.0041 $\pm$ 0.35921 & 0.1029 \\
\cline{2-5}
 & SVM & 0.0159 $\pm$ 0.00460 & 0.0120 $\pm$ 0.34777 & 0.0288 \\
\hline
2 & DT & 0.0701 $\pm$ 0.01685 & 0.0874 $\pm$ 0.35303 & 0.0850 \\
\cline{2-5}
 & SVM & 0.0162 $\pm$ 0.00669 & 0.1207 $\pm$ 0.34830 & 0.0288 \\
\hline
3 & DT & 0.0811 $\pm$ 0.02434 & 0.0394 $\pm$ 0.31475 & 0.1097 \\
\cline{2-5}
 & SVM & 0.0196 $\pm$ 0.00716 & 0.0637 $\pm$ 0.30289 & 0.0370 \\
\hline
4 & DT & 0.0636 $\pm$ 0.01976 & 0.1070 $\pm$ 0.44716 & 0.0741 \\
\cline{2-5}
 & SVM & 0.0144 $\pm$ 0.00403 & 0.1377 $\pm$ 0.43311 & 0.0274 \\
\hline
5 & DT & 0.0704 $\pm$ 0.01766 & -0.0122 $\pm$ 0.34205 & 0.0919 \\
\cline{2-5}
 & SVM & 0.0128 $\pm$ 0.00412 & 0.0376 $\pm$ 0.34765 & 0.0206 \\
\hline
6 & DT & 0.0798 $\pm$ 0.01976 & 0.0581 $\pm$ 0.41083 & 0.1001 \\
\cline{2-5}
 & SVM & 0.0169 $\pm$ 0.00535 & 0.0550 $\pm$ 0.39080 & 0.0316 \\
\hline
7 & DT & 0.0745 $\pm$ 0.01674 & 0.1047 $\pm$ 0.31367 & 0.1001 \\
\cline{2-5}
 & SVM & 0.0191 $\pm$ 0.00636 & 0.1094 $\pm$ 0.28410 & 0.0343 \\
\hline
8 & DT & 0.0663 $\pm$ 0.01578 & 0.0273 $\pm$ 0.36283 & 0.0919 \\
\cline{2-5}
 & SVM & 0.0144 $\pm$ 0.00443 & 0.0463 $\pm$ 0.36268 & 0.0219 \\
\hline
9 & DT & 0.0689 $\pm$ 0.02031 & 0.0231 $\pm$ 0.28335 & 0.0741 \\
\cline{2-5}
 & SVM & 0.0136 $\pm$ 0.00727 & 0.0265 $\pm$ 0.31730 & 0.0233 \\
\hline
10 & DT & 0.0721 $\pm$ 0.01807 & 0.0209 $\pm$ 0.36387 & 0.0808 \\
\cline{2-5}
 & SVM & 0.0152 $\pm$ 0.00435 & 0.0285 $\pm$ 0.35632 & 0.0301 \\
\hline
    \end{tabular}
\caption{\label{tab:usps-correlation} Usps: Mean bit error, mean correlation, ECOC error per fold}
\end{center}
\end{table*}

\begin{table*}[htbp]
\begin{center}
    \begin{tabular}{|c|c|c|c|c|} \hline

  Fold & Model & Mean Bit Error & Mean Correlation & ECOC Error \\
\hline
1 & DT & 0.1111 $\pm$ 0.05779 & 0.0308 $\pm$ 0.30683 & 0.1515 \\
\cline{2-5}
 & SVM & 0.1185 $\pm$ 0.06660 & 0.0077 $\pm$ 0.33212 & 0.2222 \\
\hline
2 & DT & 0.1001 $\pm$ 0.02739 & 0.0426 $\pm$ 0.31161 & 0.1212 \\
\cline{2-5}
 & SVM & 0.0790 $\pm$ 0.04514 & -0.0298 $\pm$ 0.28943 & 0.1111 \\
\hline
3 & DT & 0.1038 $\pm$ 0.03682 & 0.1067 $\pm$ 0.39916 & 0.1212 \\
\cline{2-5}
 & SVM & 0.0937 $\pm$ 0.06618 & 0.1193 $\pm$ 0.43473 & 0.1717 \\
\hline
4 & DT & 0.1414 $\pm$ 0.05448 & 0.0819 $\pm$ 0.32819 & 0.2222 \\
\cline{2-5}
 & SVM & 0.1028 $\pm$ 0.04793 & 0.0671 $\pm$ 0.38160 & 0.1919 \\
\hline
5 & DT & 0.1056 $\pm$ 0.03885 & 0.1135 $\pm$ 0.32523 & 0.1414 \\
\cline{2-5}
 & SVM & 0.1166 $\pm$ 0.06365 & 0.0227 $\pm$ 0.28839 & 0.1414 \\
\hline
6 & DT & 0.1102 $\pm$ 0.05301 & -0.0621 $\pm$ 0.32517 & 0.1414 \\
\cline{2-5}
 & SVM & 0.1093 $\pm$ 0.05855 & -0.0079 $\pm$ 0.3529 & 0.1414 \\
\hline
7 & DT & 0.1350 $\pm$ 0.04182 & 0.0115 $\pm$ 0.27311 & 0.2121 \\
\cline{2-5}
 & SVM & 0.1304 $\pm$ 0.06410 & 0.0239 $\pm$ 0.30592 & 0.2323 \\
\hline
8 & DT & 0.1010 $\pm$ 0.04476 & -0.0205 $\pm$ 0.30463 & 0.1010 \\
\cline{2-5}
 & SVM & 0.1010 $\pm$ 0.05132 & -0.0342 $\pm$ 0.35625 & 0.1515 \\
\hline
9 & DT & 0.1221 $\pm$ 0.04882 & -0.0225 $\pm$ 0.30527 & 0.1313 \\
\cline{2-5}
 & SVM & 0.1093 $\pm$ 0.05201 & 0.0533 $\pm$ 0.36505 & 0.1515 \\
\hline
10 & DT & 0.0918 $\pm$ 0.04230 & 0.0520 $\pm$ 0.30508 & 0.1010 \\
\cline{2-5}
 & SVM & 0.0955 $\pm$ 0.04484 & 0.0108 $\pm$ 0.30654 & 0.1414 \\
\hline
    \end{tabular}
\caption{\label{tab:vowel-correlation} Vowels: Mean bit error, mean correlation, ECOC error per fold}
\end{center}
\end{table*}

\begin{table*}[htbp]
\begin{center}
    \begin{tabular}{|c|c|c|c|c|} \hline

  Fold & Model & Mean Bit Error & Mean Correlation & ECOC Error \\
\hline
1 & DT & 0.0706 $\pm$ 0.00791 & 0.0061 $\pm$ 0.19321 & 0.0695 \\
\cline{2-5}
 & SVM & 0.0763 $\pm$ 0.00779 & 0.0053 $\pm$ 0.21969 & 0.1060 \\
\hline
2 & DT & 0.0644 $\pm$ 0.00818 & -0.0029 $\pm$ 0.20498 & 0.0520 \\
\cline{2-5}
 & SVM & 0.0765 $\pm$ 0.00972 & 0.0073 $\pm$ 0.20761 & 0.1060 \\
\hline
3 & DT & 0.0709 $\pm$ 0.00796 & 0.0097 $\pm$ 0.21072 & 0.0675 \\
\cline{2-5}
 & SVM & 0.0785 $\pm$ 0.00981 & 0.0082 $\pm$ 0.21389 & 0.1150 \\
\hline
4 & DT & 0.0688 $\pm$ 0.00761 & 0.0113 $\pm$ 0.19695 & 0.0635 \\
\cline{2-5}
 & SVM & 0.0732 $\pm$ 0.00988 & 0.0235 $\pm$ 0.21649 & 0.1040 \\
\hline
5 & DT & 0.0688 $\pm$ 0.00826 & 0.0129 $\pm$ 0.21209 & 0.0610 \\
\cline{2-5}
 & SVM & 0.0768 $\pm$ 0.01037 & 0.0243 $\pm$ 0.22680 & 0.1055 \\
\hline
6 & DT & 0.0652 $\pm$ 0.00815 & 0.0026 $\pm$ 0.19203 & 0.0535 \\
\cline{2-5}
 & SVM & 0.0736 $\pm$ 0.00918 & 0.0115 $\pm$ 0.20290 & 0.0985 \\
\hline
7 & DT & 0.0654 $\pm$ 0.00836 & 0.0175 $\pm$ 0.21199 & 0.0555 \\
\cline{2-5}
 & SVM & 0.0722 $\pm$ 0.01040 & 0.0027 $\pm$ 0.20112 & 0.1065 \\
\hline
8 & DT & 0.0716 $\pm$ 0.00674 & -0.0011 $\pm$ 0.19719 & 0.0620 \\
\cline{2-5}
 & SVM & 0.0761 $\pm$ 0.00909 & 0.0036 $\pm$ 0.19906 & 0.1045 \\
\hline
9 & DT & 0.0731 $\pm$ 0.00803 & -0.0034 $\pm$ 0.19219 & 0.0620 \\
\cline{2-5}
 & SVM & 0.0789 $\pm$ 0.00974 & 0.0183 $\pm$ 0.21919 & 0.1125 \\
\hline
10 & DT & 0.0668 $\pm$ 0.00802 & 0.0049 $\pm$ 0.22051 & 0.0665 \\
\cline{2-5}
 & SVM & 0.0739 $\pm$ 0.00927 & 0.0112 $\pm$ 0.21925 & 0.1010 \\
\hline
    \end{tabular}
\caption{\label{tab:letters-correlation} Letters: Mean bit error, mean correlation, ECOC error per fold}
\end{center}
\end{table*}

\begin{table*}[htbp]
\begin{center}
    \begin{tabular}{|c|c|c|c|} \hline

  Fold & Mean Bit Error & Mean Correlation & ECOC Error \\
\hline
1 & 0.0141 $\pm$ 0.00444 & 0.2054 $\pm$ 0.07427 & 0.0203 \\
\hline
2 & 0.0177 $\pm$ 0.00555 & 0.2354 $\pm$ 0.06582 & 0.0242 \\
\hline
3 & 0.0170 $\pm$ 0.00560 & 0.2417 $\pm$ 0.06188 & 0.0247 \\
\hline
4 & 0.0168 $\pm$ 0.00581 & 0.2253 $\pm$ 0.07316 & 0.0225 \\
\hline
5 & 0.0154 $\pm$ 0.00580 & 0.2153 $\pm$ 0.08254 & 0.0218 \\
\hline
6 & 0.0161 $\pm$ 0.00561 & 0.2326 $\pm$ 0.07463 & 0.0223 \\
\hline
7 & 0.0174 $\pm$ 0.00587 & 0.2174 $\pm$ 0.08201 & 0.0248 \\
\hline
8 & 0.0163 $\pm$ 0.00566 & 0.2131 $\pm$ 0.06321 & 0.0232 \\
\hline
9 & 0.0150 $\pm$ 0.00516 & 0.2035 $\pm$ 0.07747 & 0.0203 \\
\hline
10 & 0.0161 $\pm$ 0.00548 & 0.2145 $\pm$ 0.06719 & 0.0217 \\
\hline
    \end{tabular}
\caption{\label{tab:cifar-correlation} CIFAR-10: Mean bit error, mean correlation, ECOC error per fold}
\end{center}
\end{table*}

\begin{table*}[htbp]
\begin{center}
    \begin{tabular}{|c|c|c|c|} \hline

  Fold & Mean Bit Error & Mean Correlation & ECOC Error \\
\hline
1 & 0.0082 $\pm$ 0.00186 & 0.2153 $\pm$ 0.05684 & 0.0116 \\
\hline
2 & 0.0089 $\pm$ 0.00223 & 0.1698 $\pm$ 0.04895 & 0.0109 \\
\hline
3 & 0.0083 $\pm$ 0.00160 & 0.1922 $\pm$ 0.06399 & 0.0108 \\
\hline
4 & 0.0081 $\pm$ 0.00178 & 0.1783 $\pm$ 0.05527 & 0.0107 \\
\hline
5 & 0.0087 $\pm$ 0.00194 & 0.1644 $\pm$ 0.04570 & 0.0108 \\
\hline
6 & 0.0082 $\pm$ 0.00156 & 0.1766 $\pm$ 0.05912 & 0.0092 \\
\hline
7 & 0.0094 $\pm$ 0.00210 & 0.2134 $\pm$ 0.05468 & 0.0124 \\
\hline
8 & 0.0088 $\pm$ 0.00162 & 0.2033 $\pm$ 0.04908 & 0.0125 \\
\hline
9 & 0.0081 $\pm$ 0.00162 & 0.1723 $\pm$ 0.03762 & 0.0097 \\
\hline
10 & 0.0091 $\pm$ 0.00222 & 0.1746 $\pm$ 0.05345 & 0.0121 \\
\hline
    \end{tabular}
\caption{\label{tab:svhn-correlation} SVHN: Mean bit error, mean correlation, ECOC error per fold}
\end{center}
\end{table*}

\end{document}